\documentclass[10pt,journal,compsoc]{IEEEtran}

\usepackage{cite}
\usepackage[pdftex]{graphicx}
\usepackage{amsmath}
\usepackage{amssymb}
\usepackage{amsfonts}
\usepackage{algorithmic}
\usepackage{array}
\usepackage{color}

\usepackage{multirow}
\usepackage{multicol}
\usepackage{hhline}
\usepackage{booktabs}
\usepackage{tabularx}
\usepackage{anyfontsize}
\usepackage{threeparttable}
\usepackage{ragged2e}
\usepackage{makecell}
\usepackage[pagebackref,breaklinks,colorlinks,bookmarks=false,citecolor=blue,linkcolor=blue]{hyperref}
\usepackage{paralist}
\usepackage[table]{xcolor}
\definecolor{lightgray}{gray}{0.9}

\renewcommand{\raggedright}{\leftskip=0pt \rightskip=0pt plus 0cm}

\newcommand{\ra}[1]{\renewcommand{\arraystretch}{#1}}

\usepackage{mathtools}
\DeclarePairedDelimiterX{\infdivx}[2]{(}{)}{%
  #1\;\delimsize\|\;#2%
}
\newcommand{\infdiv}{D_{KL}\infdivx}

\DeclareMathOperator*{\argmaxA}{arg\,max}

\begin{document}
\title{GAN-based Facial Attribute Manipulation
}
%
%

\author{
Yunfan~Liu,~Qi~Li,~Qiyao~Deng,~Zhenan~Sun,~and~Ming-Hsuan~Yang%
\IEEEcompsocitemizethanks{
\IEEEcompsocthanksitem Yunfan Liu and Zhenan Sun are with the School of Artificial Intelligence, University of Chinese Academy of Sciences, Beijing 100049, China, and are also with the National Laboratory of Pattern Recognition, Center for Research on Intelligent Perception and Computing, Institute of Automation, Chinese Academy of Sciences, Beijing, 100190, China (e-mail: yunfan.liu@cripac.ia.ac.cn; znsun@nlpr.ia.ac.cn).%
\IEEEcompsocthanksitem Qi Li and Qiyao Deng are with the Center for Research on Intelligent Perception and Computing, Institute of Automation, Chinese Academy of Sciences, Beijing, 100190, China (e-mail: qli@nlpr.ia.ac.cn; dengqiyao@cripac.ia.ac.cn).%
\IEEEcompsocthanksitem Ming-Hsuan Yang is with the Department of Computer Science and Engineering, University of California, Merced, CA, 95340, USA (email: mhyang@ucmerced.edu).%
}
}

%

\IEEEtitleabstractindextext{%
\raggedright{
\begin{abstract}
Facial Attribute Manipulation (FAM) aims to aesthetically modify a given face image to render desired attributes, which has received significant attention due to its broad practical applications ranging from digital entertainment to biometric forensics.
In the last decade, with the remarkable success of Generative Adversarial Networks (GANs) in synthesizing realistic images, numerous GAN-based models have been proposed to solve FAM with various problem formulation approaches and guiding information representations.
This paper presents a comprehensive survey of GAN-based FAM methods with a focus on summarizing their principal motivations and technical details.
The main contents of this survey include: (i) an introduction to the research background and basic concepts related to FAM, (ii) a systematic review of GAN-based FAM methods in three main categories, and (iii) an in-depth discussion of important properties of FAM methods, open issues, and future research directions.
This survey not only builds a good starting point for researchers new to this field but also serves as a reference for the vision community.
\end{abstract}}

}

\maketitle
\IEEEdisplaynontitleabstractindextext
\IEEEpeerreviewmaketitle

\IEEEraisesectionheading{\section{Introduction}\label{sec:introduction}}

%
%
%
%

\IEEEPARstart{F}{acial} attributes describe the semantic properties of human faces in an interpretable manner.
As soft biometrics presenting the classifiable visual characteristic of faces with rich individual variations, facial attributes have been playing a vital role in research on many computer vision tasks, such as face recognition~\cite{hu2017attribute,liu2018deep,huang2019deep,jiang2019robust} and face retrieval~\cite{han2017heterogeneous,zaeemzadeh2021face,fang2021attribute}.
With the development of deep generative models, 
\textnormal{facial attribute manipulation (FAM)} has attracted growing research attention due to its profound theoretical significance and broad practical applications.

\begin{figure}[t]
\begin{center}
\includegraphics[width=0.95\linewidth]{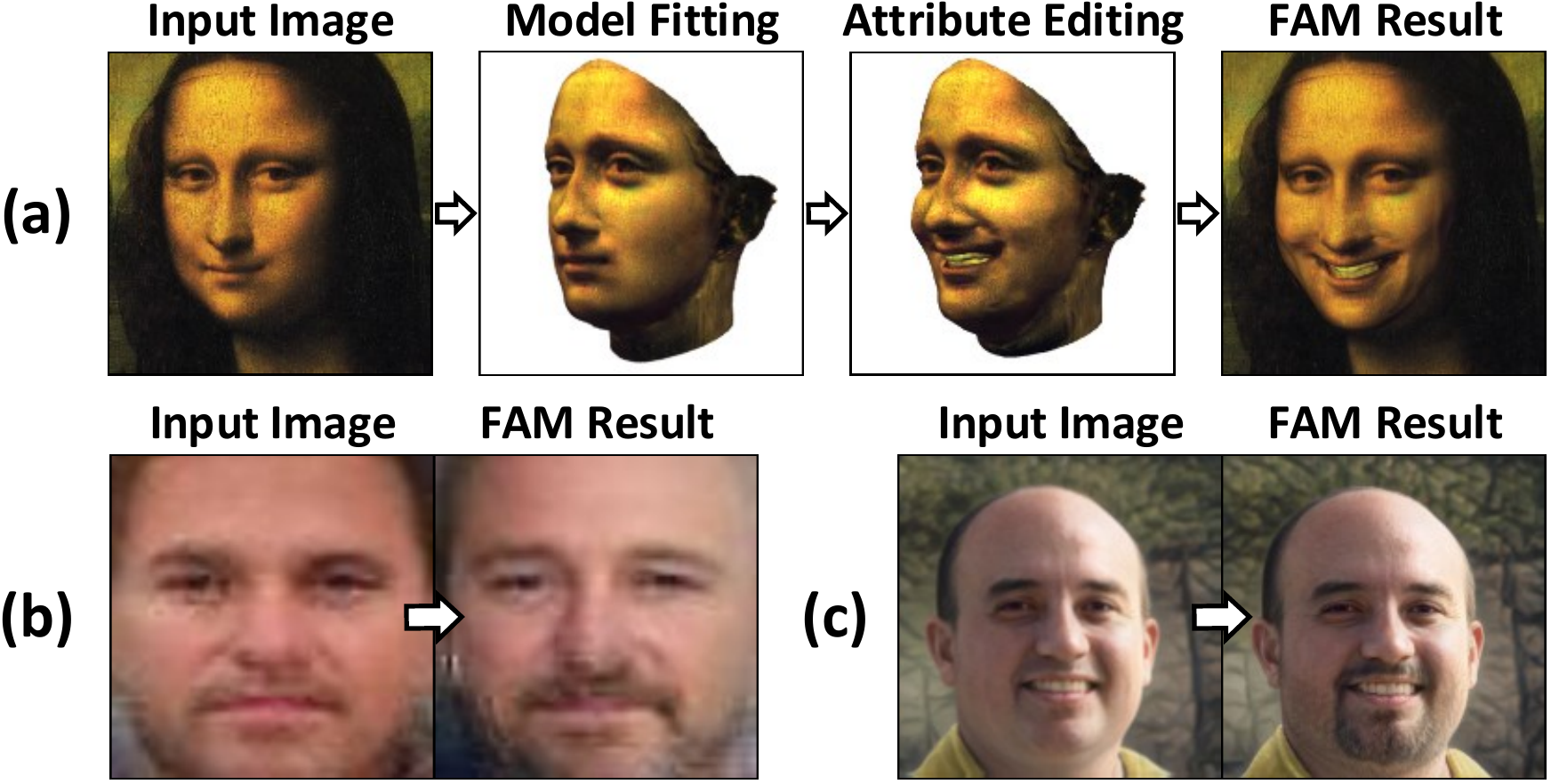}
\end{center}
\vspace{-0.2cm}
\caption{Sample results obtained by (a) model-driven FAM methods~\cite{blanz2003reanimating} (Smiling), (b) FAM methods built on deep networks~\cite{nhan2017temporal} (Age), and (c) GAN-based FAM methods~\cite{yao2021latent} (Beard). Images are directly obtained from the corresponding paper for illustrative purposes.
\vspace{-0.2cm}
}
\label{fig:intro_FAM_samples}
\end{figure}

Existing FAM methods aim to modify input face images to render target attributes while ensuring the visual quality of manipulation results (see Fig.~\ref{fig:intro_FAM_samples}).
Early research on FAM can be traced back to the 1990s~\cite{wu1994plastic,rowland1995manipulating}, where the modeling of faces largely involves complex prior knowledge specific to the target task.
As a class of parameterized models describing the principal variations of face shape, 3D Morphable Models (3DMMs)~\cite{blanz1999morphable,paysan20093d,cao2013facewarehouse} are widely used in many studies to manipulate attributes related to facial geometry (e.g., pose~\cite{hassner2013viewing,taigman2014deepface,hassner2015effective} and expression~\cite{blanz2003reanimating,cao2014displaced,zhu2015high}).
Although FAM can be performed by these \textnormal{model-driven} approaches, the algorithm pipelines are usually complicated, and synthesized results are of low visual quality (see Fig~\ref{fig:intro_FAM_samples}(a)).
To solve this problem,  subsequent studies~\cite{zhu2016face,thies2016face2face,wang2016recurrent,upchurch2017deep,nhan2016longitudinal,nhan2017temporal} adopt \textnormal{data-driven} deep models for feature extraction and image translation, which can generate high-quality images in an end-to-end manner.

In the last decade, the development of Generative Adversarial Networks (GANs)~\cite{goodfellow2014generative} have made significant advances in improving the visual quality of synthesized, images~\cite{radford2016unsupervised,karras2017progressive,karras2019style,karras2020analyzing,karras2021alias}, and GAN-based FAM has received much attention from the community of computer vision.
Due to the intrinsic complexity of face images and semantic variations of facial attributes, FAM is one of the most important applications of GANs for studying controllable and diverse image translation.
It not only serves as a common benchmark for general-purpose image translation models based on GANs~\cite{he2019attgan,choi2018stargan,shen2020interpreting}, but also motivates numerous solutions based on various problem settings~\cite{deng2020disentangled,kowalski2020config,shu2017neural,gu2019mask,lee2020maskgan,zhu2020sean}.
Moreover, FAM also has great practical significance as it is closely related to many real-world applications, such as digital entertainment (editing photos posted on social media), face recognition (normalizing pose and expression), and information forensics (augmenting training data for DeepFake detection).

In this paper, we present a comprehensive survey of existing GAN-based FAM methods with a focus on discussing the motivations and features, as well as summarizing the similarities and differences to give a whole picture of representative approaches in this field.
Moreover, we also identify challenges and open issues of existing studies and forecast promising future research directions.
%
Although some surveys~\cite{zheng2020survey} and~\cite{pang2021image}
have also discussed GAN-based FAM methods, they do not include those built on pre-trained generative priors, which have been extensively studied in recent years. 
On the other hand, the review~\cite{xia2021gan} focuses on GAN inversion methods, and FAM is only considered as one of the downstream applications.
Moreover, FAM is also discussed in~\cite{wang2020survey} and~\cite{tolosana2020deepfakes}, but the main themes are on data augmentation and DeepFake detection, respectively.

The rest of this article is organized as follows.
In Section~\ref{sec:pre}, we define important concepts in FAM and the scope of this survey, and then categorize existing GAN-based FAM methods in terms of methodology.
To conduct a self-contained review of literature, the basic formulation of GAN models, the commonly-used datasets, and metrics for quantitative evaluation are introduced in Section~\ref{sec:basic}.
Representative GAN-based FAM methods are then systematically reviewed in Section~\ref{sec:IDT_FAM}, Section~\ref{sec:FSD_FAM}, and Section~\ref{sec:LSN_FAM}.
In Section~\ref{sec:property}, we summarize the important properties of FAM and how they are approached by GAN-based methods.
We discuss the challenges and open issues in FAM, as well as promising future research directions in Section~\ref{sec:challenge_future}.

\begin{figure}[t]
\begin{center}
\includegraphics[width=0.85\linewidth]{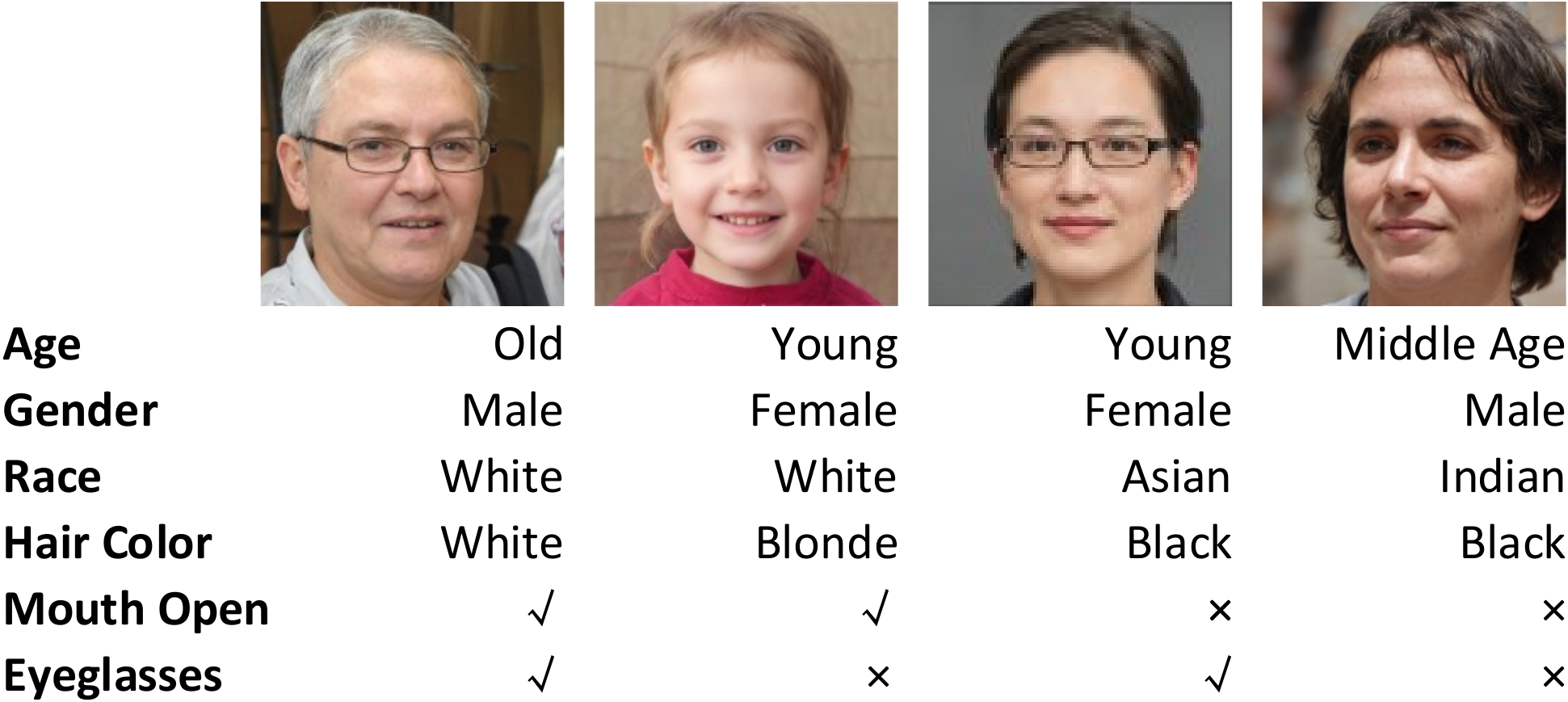}
\end{center}
\vspace{-0.2cm}
\caption{Illustration of sample facial attributes. 
Note that the same class of a certain attribute may have large individual variations in visual appearance (e.g., eyeglasses of different styles).
}
\vspace{-0.2cm}
\label{fig:pre_FA_concept}
\end{figure}

\section{Preliminaries}\label{sec:pre}

\subsection{Facial Attribute Manipulation}

%
In recent years, numerous GAN models have been developed for \textnormal{image-to-image translation (I2I)}~\cite{pang2021image}, which is the broadest concept among all topics related to image manipulation.
%
On the other hand, \textnormal{face image translation} aims to change holistic appearance, including identity swapping~\cite{li2020advancing,zhu2021one,gao2021information} and make-up transfer~\cite{gu2019ladn,chen2019beautyglow,jiang2020psgan,deng2021spatially}.
\textnormal{Facial attribute manipulation}, the research topic of this survey, further constrains the manipulated semantics to facial attributes.

\begin{figure}[t]
\begin{center}
\includegraphics[width=0.95\linewidth]{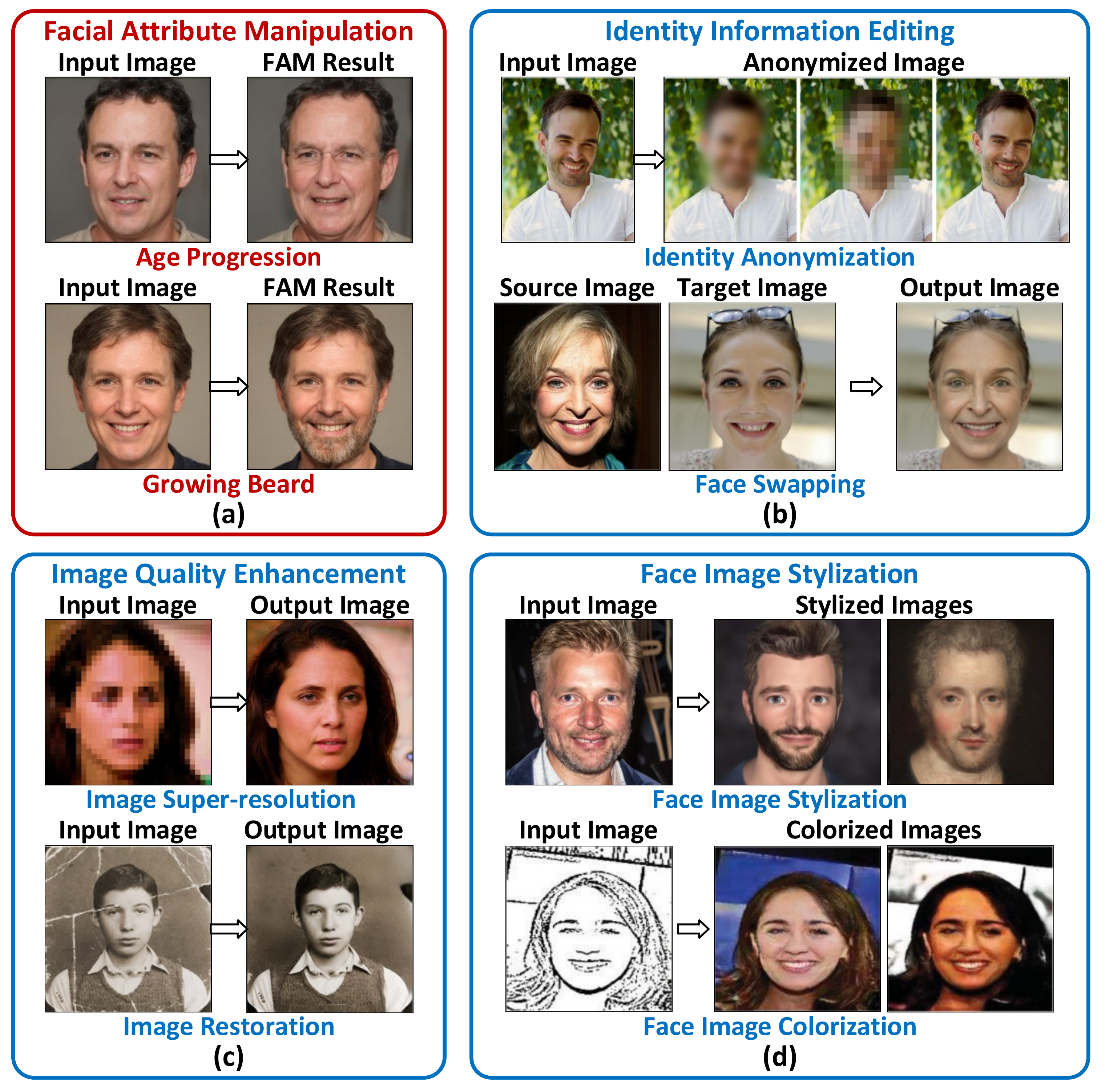}
\end{center}
\vspace{-0.2cm}
\caption{Sample results of (a) Facial Attribute Manipulation and other face image translation tasks, including (b) Identity Information Editing (e.g., Identity Anonymization~\cite{meden2021privacy} and Face Swapping~\cite{zhu2021one}); (c) Image Quality Improvement (e.g., Face Image Super-resolution\cite{menon2020pulse} and Face Image Restoration~\cite{wan2022old}); and (d) Face Image Stylization (e.g., Face Image Stylization~\cite{song2021agilegan} and Face Image Colorization~\cite{lee2020reference})).
Images are directly obtained from the correspondingly referenced paper.
}
\vspace{-0.2cm}
\label{fig:pre_topic_dist}
\end{figure}

In this paper, we define \textnormal{facial attributes} as the inherent properties of human faces which are categorical and interpretable.
They describe generic visual characteristics of facial components (including hair and accessories) or soft-biometrics (e.g., age, gender, and race), which not only can be categorized into discrete classes but also contain large individual variation in terms of textural details (see Fig.~\ref{fig:pre_FA_concept}).
As such, FAM can be clearly distinguished from the following face image translation tasks:
\begin{enumerate}
  \item \textnormal{Identity information editing}: In this paper, identity information is not considered as a facial attribute since it does not present an interpretable or categorical feature of human faces. 
  Therefore, related tasks such as identity anonymization~\cite{maximov2020ciagan,chen2021perceptual,cao2021personalized} and face swapping~\cite{li2020advancing,zhu2021one,gao2021information} are beyond the scope of FAM.
  \item \textnormal{Image quality enhancement}: Although the prior knowledge of facial biometrics is widely incorporated when improving the quality of portrait images, such as in image super-resolution~\cite{yu2018super,chen2018fsrnet,yu2019semantic} or image restoration~\cite{wan2022old}, images are translated in terms of the quality of textural details or overall perception, which are not considered as facial attributes.
  \item \textnormal{Face image stylization}: Similar to image quality, artistic styles are not biological properties specific to face images, and thus can not be treated as facial attributes. Therefore, face image cartoonization~\cite{kim2019u,pinkney2020resolution,song2021agilegan} and colorization~\cite{he2019adversarial,lee2020reference} are out of the scope of this survey.
\end{enumerate}
Fig.~\ref{fig:pre_topic_dist} presents sample results and illustrates the difference between FAM and other face image translation tasks.

\subsection{Taxonomy}
A typical GAN model consists of two networks, a generator $G$ mapping an input variable $\boldsymbol{z}\sim p_z$ to the output image ($p_z$ denotes the prior distribution), and a discriminator $D$ distinguishing generated images $G(\boldsymbol{z})$ from real ones $\boldsymbol{x}$.
These two modules are adversarially trained to encourage $G$ to produce visually plausible output images.


\begin{figure}[t]
\begin{center}
\includegraphics[width=0.90\linewidth]{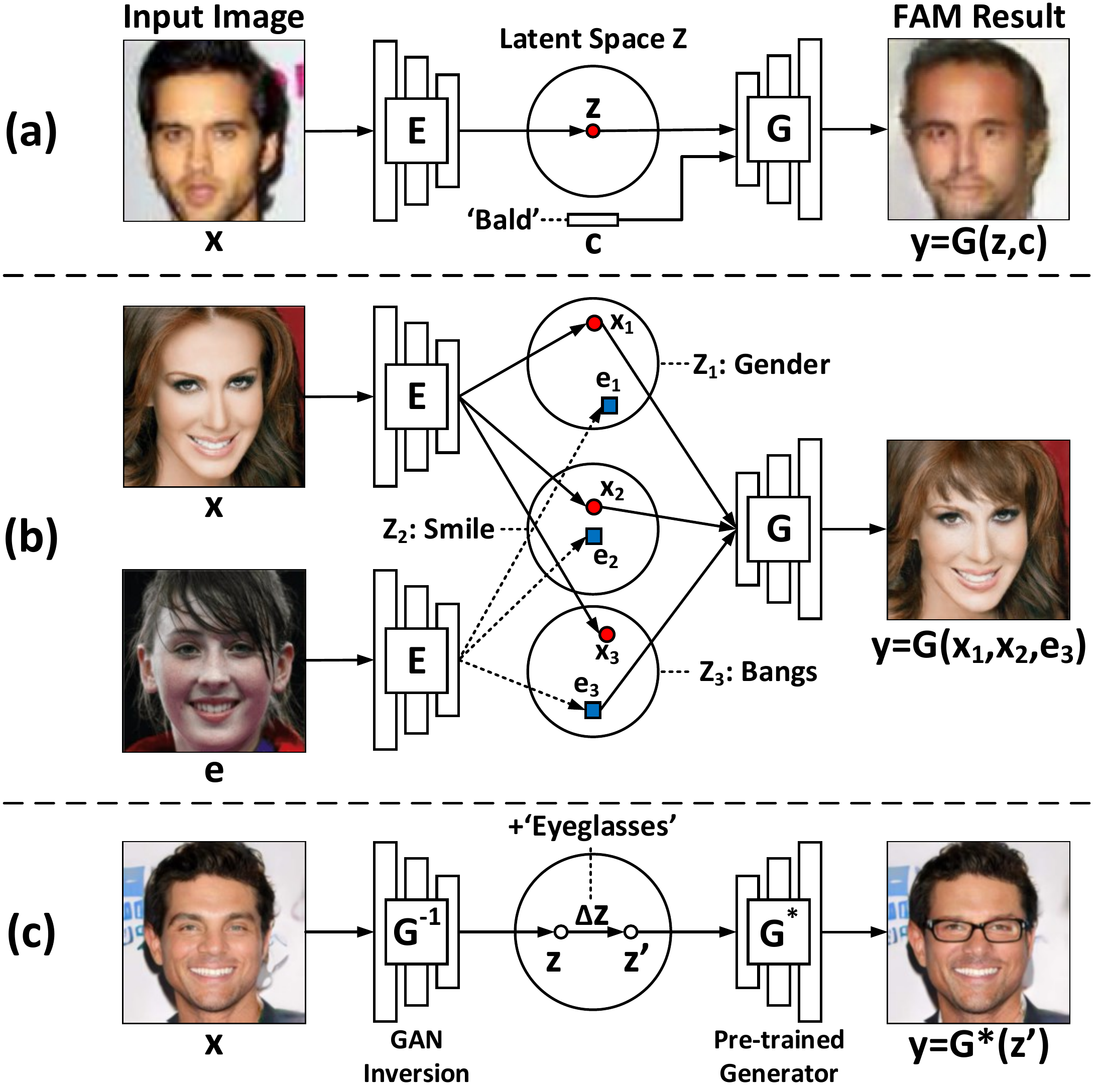}
\end{center}
\vspace{-0.2cm}
\caption{Framework of (a) image domain translation based FAM methods (Bald~\cite{perarnau2016invertible}), (b) facial semantic decomposition based FAM methods (Bangs~\cite{xiao2018elegant}), where $\boldsymbol{e}$ denotes the exemplar image, and (c) latent space navigation based FAM methods (Eyeglasses~\cite{shen2020interpreting}).
}
\vspace{-0.2cm}
\label{fig:intro_FAM_teaser}
\end{figure}

For a GAN-based FAM method where the input is a face image $\boldsymbol{x}$, an encoding process $E$ is usually required to project $\boldsymbol{x}$ into a latent space $Z$ and obtain its embedded representation $\boldsymbol{z}=E(\boldsymbol{x})\in Z$ for further manipulation.
Based on \textnormal{how $\boldsymbol{z}$ is obtained and manipulated to render the desired attribute change}, we broadly categorize GAN-based FAM methods into three main groups:  {\textnormal{image domain translation}} based methods, {\textnormal{facial semantic decomposition}} based methods, and {\textnormal{latent space navigation}} based approaches (Fig.~\ref{fig:intro_FAM_teaser}).

\vspace{1mm}
\noindent \textbf{Image domain translation based methods.} These approaches consider FAM as an image-to-image translation (I2I) problem, where face images are grouped into distinctive categories (i.e., image domains) according to their attributes, and domain-level mapping functions are learned to perform image translation.
Early work in this category mainly focuses on manipulating only one facial attribute with a single model~\cite{isola2017image,zhu2017unpaired,shen2017learning}, and subsequent schemes propose to scale to multi-attribute editing by incorporating various kinds of conditional information (denoted as $\boldsymbol{c}$ in Fig.~\ref{fig:intro_FAM_teaser}(a))~\cite{perarnau2016invertible,he2019attgan,choi2018stargan,wu2019relgan,liu2019stgan}.

Instead of computing a joint latent representation $\boldsymbol{z}$ for the input image $\boldsymbol{x}$, \textnormal{facial semantic decomposition based FAM methods}~\cite{lee2018diverse,huang2018multimodal,cho2019image,park2020swapping,lee2020drit++,choi2020stargan,li2021image,xiao2018dna,xiao2018elegant,shoshan2021gan,deng2020disentangled,kowalski2020config,shu2017neural,gu2019mask,zhu2020sean,lee2020maskgan} encode $\boldsymbol{x}$ into separate latent spaces, where the embedding in each space is responsible for controlling different image semantics of $\boldsymbol{x}$. 
Compared to domain-level image translation, this class of methods successfully captures the intra-domain image variation, i.e., the difference in detailed style information of an attribute (e.g., the appearance of bangs in Fig.~\ref{fig:intro_FAM_teaser}(b)), which enables generating diverse FAM results with fine-grained controllability and high flexibility.

Although realistic FAM results can be generated by the above-mentioned methods, these approaches can hardly be directly applied to high-resolution (HR) face images due to the extremely increased computational cost of training.
Therefore, \textnormal{latent space navigation based approaches}~\cite{shen2020interpreting,yang2021discovering,tewari2020stylerig,abdal2021styleflow,shen2021closed} 
address the FAM problem by making the embedding $\boldsymbol{z}=G^{-1}(\boldsymbol{x})$ ($G^{-1}$ denotes the GAN inversion operation~\cite{xia2021gan}) travel through the latent space $Z$ of a pre-trained generator $G^\ast$, such that the decoding result of the translated latent code $\boldsymbol{z}'$, i.e., $G^\ast(\boldsymbol{z}')$, can present the desired facial attributes.
Latent space navigation based FAM methods have attracted remarkable research attention since the advent of large-scale GAN models (e.g., StyleGANs~\cite{karras2019style,karras2020analyzing,karras2021alias}), as the generator network can be efficiently scaled to HR images ($1024\times 1024$) without training from scratch.

\section{Formulation, Dataset, and Metrics}\label{sec:basic}
For a self-contained literature review on FAM, in this section, we introduce basic formulations of GANs, datasets in GAN-based FAM methods, and commonly used metrics for quantitative evaluation.

\subsection{Basic Formulations of GANs}

\begin{figure}[t]
\begin{center}
\includegraphics[width=0.90\linewidth]{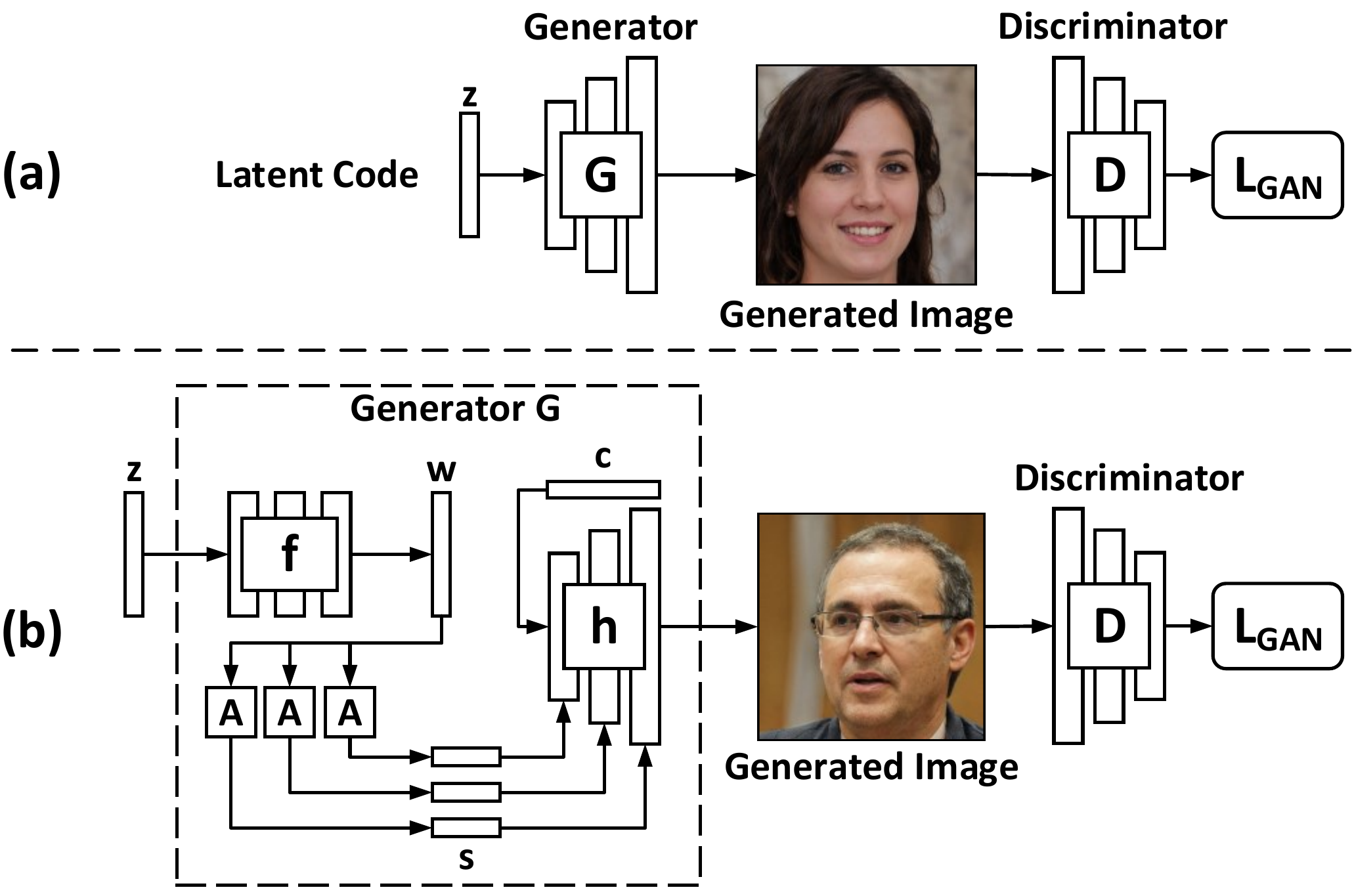}
\end{center}
\vspace{-0.2cm}
\caption{Overview of the framework of (a) GAN model with a typical generator and (b) GAN model with the style-based generator.
}
\vspace{-0.2cm}
\label{fig:GAN_model}
\end{figure}

GANs are a class of deep neural networks that learns to estimate the mapping $\mathcal{G}:Z\rightarrow X$ with a generator network $G$, where $Z$ is a latent space with a tractable prior $p_{z}$ and $X$ is the target space with an intractable data distribution $p_{x}$.
Moreover, a discriminator $D$ is adopted to distinguish $p_{g}$ from $p_{x}$, where $p_{g}$ denotes the distribution of the output of $G$.
The parameters of $G$ and $D$ are optimized alternatively via an adversarial process by
\begin{align}
\begin{split}
\min_{G} \max_{D} L_{GAN} & = \mathbb{E}_{\boldsymbol{x}\sim p_{x}(\boldsymbol{x})}[\log D(\boldsymbol{x})] \\
                              & + \mathbb{E}_{\boldsymbol{z}\sim p_{z}(\boldsymbol{z})}[\log (1-D(G(\boldsymbol{z})))]
\end{split}
\end{align}
Subsequent to the original GAN~\cite{goodfellow2014generative}, many methods have been proposed to stabilize the training process and improve the quality of generated images by developing advanced objective functions~\cite{mao2017least,arjovsky2017wasserstein,gulrajani2017improved}. and network structures~\cite{radford2016unsupervised,karras2017progressive}.
%
As shown in Fig.~\ref{fig:GAN_model}(a), the generator of these GAN models directly takes the latent code $\boldsymbol{z}$ as input to the stacked convolutional layers and produces the output image, which we term as \textnormal{traditional generators}.

In contrast, the \textnormal{style-based generators}~\cite{karras2019style,karras2020analyzing,karras2021alias} take a learnable constant tensor $\boldsymbol{c}$ as the input to the synthesis network $h$ (see Fig.~\ref{fig:GAN_model}(b)).
To control the semantics of generation results, a latent code $\boldsymbol{z}\in Z$ is firstly projected into an intermediate latent space $W$ via a mapping network $f$, i.e., $\boldsymbol{w}=f(\boldsymbol{z})\in W$.
Afterwards, $\boldsymbol{w}$ is replicated by $N$ times and fed into separate affine transformation networks $A_i$ $(i=1,2,\ldots,N)$ to compute coefficients $\boldsymbol{s_i}=A_i(\boldsymbol{w})$ (i.e., style codes), which are used to modulate feature maps in $h$ via adaptive instance normalization (AdaIN) modules~\cite{huang2017arbitrary}. 
Notably, the space spanned by channel-wise style parameters is often referred to as the style space $S$~\cite{liu2020style,wu2021stylespace}, and some GAN inversion methods~\cite{abdal2020image2stylegan++} propose to approximate the input image by reconstructing with different $\boldsymbol{w}$ codes, which form the $W^+$ space.

\subsection{Facial Attribute Datasets}

Numerous datasets have been used for FAM:
\begin{compactitem}
\item Labeled Faces in the Wild-attribute (LFWA)~\cite{liu2015deep} is created by annotating 40 facial attributes with binary labels of images in LFW~\cite{huang2008labeled}. It contains 13,233 face images collected from the Internet with 5,729 identities.

\item Radboud Faces Database (RaFD)~\cite{langner2010presentation} contains 4,824 images of 67 subjects captured in controlled environments with simple background.
Each subject is asked to make 8 facial expressions with 3 different gaze directions, and each face is captured from 3 different angles.

\item 
CelebFaces Attributes (CelebA)~\cite{liu2015deep} is the most widely used large-scale face dataset in GAN-based FAM studies due to its diversity of attribute labels and large variation of image content.
It provides binary labels of 40 facial attributes for 202,599 celebrity face images with 10,177 identities in CelebFaces~\cite{sun2014deep} (see Table~\ref{table:facial_attributes}).

\item CelebA-HQ~\cite{karras2017progressive} is the high-quality version of CelebA~\cite{liu2015deep} consisting of 30,000 images at $1024\times 1024$ resolution with the same attribute label. 
It is created by enhancing and cropping the face regions of all images in CelebA and selecting samples with the best visual quality.

\item 
CelebAMask-HQ~\cite{lee2020maskgan} further provides the parsing map of images in CelebA-HQ down-sampled to $512\times 512$, where pixel-level annotation of 19 classes, including facial components and accessories, are provided.

\item Flickr-Faces-HQ (FFHQ)~\cite{karras2019style} contains 70,000 face images at $1024\times 1024$ resolution, which were originally crawled from Flickr, manually checked to discard low-quality samples, and then normalized with dlib~\cite{dlib09}.
\end{compactitem}



\begin{table}[t]
\caption{Categorization of 40 facial attributes labeled in CelebA, CelebA-HQ, and CelebAMask-HQ.}
\vspace{-0.2cm}
\centering
\resizebox{1.0\columnwidth}{!}{%
\ra{1.3}
\begin{tabular}{cl}
\hline
Component & \multicolumn{1}{c}{Attributes} \\
\hline
\cellcolor[gray]{0.9}  & \cellcolor[gray]{0.9}Black\_Hair, Blonde\_Hair, Brown\_Hair, Gray\_Hair, Wavy\_Hair, \\
\multirow{-2}{*}{\cellcolor[gray]{0.9}Hair}  & \cellcolor[gray]{0.9}Straight\_Hair, Receding\_Hairline, Bald, Bangs, Sideburns \\
\multirow{2}{*}{Eyes}  & Eyeglasses, Bushy\_Eyebrows, Narrow\_Eyes, Bags\_Under\_Eyes \\
                       & Arched\_Eyebrows \\
\cellcolor[gray]{0.9}  & \cellcolor[gray]{0.9}Mouth\_Slightly\_Open, Goatee, Mustache, Smiling, Big\_Lips, \\
\multirow{-2}{*}{\cellcolor[gray]{0.9}Mouth} & \cellcolor[gray]{0.9}No\_Beard, Wearing\_Lipstick \\
Nose  & Big\_Nose, Pointy\_Nose \\
\cellcolor[gray]{0.9}  & \cellcolor[gray]{0.9}Oval\_Face, Double\_Chin, Rosy\_Cheeks, Heavy\_Makeup, Pale\_Skin \\
\multirow{-2}{*}{\cellcolor[gray]{0.9}Skin}  & \cellcolor[gray]{0.9}High\_Cheekbones  \\
Accessory & Wearing\_Earrings, Wearing\_Hat, Wearing\_Necklace, Wearing\_Necktie\\
\cellcolor[gray]{0.9}General & \cellcolor[gray]{0.9}Male, Young, Chubby, Attractive, Blurry, 5\_o\_Clock\_Shadow \\
\hline
\end{tabular}
}
\label{table:facial_attributes}
\end{table}

\subsection{Evaluation Metrics}

We summarize evaluation metrics widely used  for FAM in three aspects: \textnormal{realism} of manipulated images,  \textnormal{accuracy} of attribute translation, and \textnormal{consistency} of image semantics.

\subsubsection{Realism of Manipulated Images}

The \textnormal{Inception Score} (IS)~\cite{salimans2016improved} has been used as an alternative to human annotators for measuring both the \textnormal{diversity} and \textnormal{interpretability} of images generated by GANs.
These goals can be jointly measured by the following objective function
\begin{equation}
\text{IS} = \exp{(\mathbb{E}_{\boldsymbol{x}\sim p_g}\infdiv{p(\boldsymbol{y}|\boldsymbol{x})}{p(\boldsymbol{y})})}
\label{eq:IS}
\end{equation}
where $D_{KL}$ denotes the KL-Divergence, and $p_{\boldsymbol{x}\sim p_g}(\boldsymbol{y}|\boldsymbol{x})$ is the conditional label distribution computed by a pre-trained Inception-v3 network~\cite{szegedy2016rethinking}.

The \textnormal{Fr$\boldsymbol{\acute{e}}$chet Inception Distance} (FID)~\cite{heusel2017gans} measures the discrepancy between $p_g$ and $p_x$, while IS is computed only based on $p_g$.
Concretely, FID computes the Fr$\acute{e}$chet Distance between two Gaussian distributions, $\mathcal{N}(\boldsymbol{\mu_g}, \boldsymbol{\Sigma_g})$ and $\mathcal{N}(\boldsymbol{\mu_r}, \boldsymbol{\Sigma_r})$, which denote the distribution of Inception representations~\cite{szegedy2016rethinking} extracted from generated images and real training images.
The FID metric is computed by
\begin{equation}
\text{FID} = \Vert\boldsymbol{\mu_g}-\boldsymbol{\mu_r}\Vert^2 + \text{Tr}(\boldsymbol{\Sigma_g}+\boldsymbol{\Sigma_r}-2(\boldsymbol{\Sigma_g}\boldsymbol{\Sigma_r})^{\frac{1}{2}})
\label{eq:FID}
\end{equation}

The \textnormal{Kernel Inception Distance} (KID)~\cite{binkowski2018demystifying} measures the maximum mean discrepancy (MMD) between the Inception embedding~\cite{szegedy2016rethinking} of two images (denoted as $\boldsymbol{x}$ and $\boldsymbol{y}$) as 
\begin{equation}
\text{KID} = (\frac{1}{d}\boldsymbol{x}^{\top}\boldsymbol{y}+1)^3
\label{eq:KID}
\end{equation}
Different from FID, KID does not assume the distribution of activations to be parametric, and it also compares skewness as an addition to mean and variance.

The \textnormal{Learned Perceptual Image Path Similarity} (LPIPS)~\cite{zhang2018unreasonable} measures the perceptual similarity between two images using their deep embeddings.
Given two image patches, $\boldsymbol{x}$ and $\boldsymbol{x_0}$, features at $L$ layers are extracted and then normalized to unit length, which are denoted as $\hat{\boldsymbol{y}}^l, \hat{\boldsymbol{y}}^l_0\in \mathbb{R}^{H_l\times W_l \times C_l}$ for the $l$-th layer.
The LPIPS score is computed as
\begin{equation}
\text{LPIPS}(\boldsymbol{x}, \boldsymbol{x_0})=\sum_{l}\frac{1}{H_l W_l}\sum_{h,w}\Vert \boldsymbol{w}^l\cdot (\hat{\boldsymbol{y}_{hw}}^l - \hat{\boldsymbol{y}}^l_{0hw})\Vert_2^2
\end{equation}
where $\boldsymbol{w}^l\in \mathbb{R}^{C_l}$ are learnable channel-wise scaling weights.

\begin{table*}[t]
\caption{Summary of the characteristics of FAM methods based on image domain translation. 
Guidance Type includes Abs. and Rel., which refer to Absolute and Relative, respectively.}
\vspace{-0.2cm}
\rowcolors{1}{}{lightgray}
\centering
\ra{1.3}
\resizebox{2.0\columnwidth}{!}{%
\begin{tabular}{cccccccc}
\hline
Method Name                               & Publication  & Model Structure & Scalability & Guidance Type & Diversity & Resolution & Quantitative Metrics\\
\hline
CycleGAN~\cite{zhu2017unpaired}           & ICCV 2017    & Cyclic    & Two-domain   & -     & Unimodal    & $256\times 256$ & US \\
DualGAN~\cite{yi2017dualgan}              & ICCV 2017    & Cyclic    & Two-domain   & -     & Unimodal    & $256\times 256$ & US \\
DiscoGAN~\cite{kim2017learning}           & ICML 2017    & Cyclic    & Two-domain   & -     & Unimodal    & $64\times 64$   & RMSE of Pose \\
UNIT~\cite{liu2017unsupervised}           & NeurIPS 2017 & Cyclic    & Two-domain   & -     & Unimodal    & $256\times 256$ & TARR, NAPR \\
AugCycleGAN~\cite{almahairi2018augmented} & ICML 2017    & Cyclic    & Two-domain   & -     & Multi-modal & $128\times 128$ & LPIPS, Precision, normalized Discounted Cumulative Gain \\
ResGAN~\cite{shen2017learning}            & CVPR 2017    & Cyclic    & Two-domain   & -     & Unimodal    & $128\times 128$ & Error of Facial Landmarks \\
ACL-GAN~\cite{zhao2020unpaired}           & ECCV 2020    & Cyclic    & Two-domain   & -     & Multi-modal & $256\times 256$ & FID, KID \\
AttentionGAN~\cite{tang2021attentiongan}  & TNNLS 2021   & Cyclic    & Two-domain   & -     & Unimodal    & $256\times 256$ & FID, KID, PSNR, US \\
Pix2pix~\cite{isola2017image}             & CVPR 2017    & One-sided & Two-domain   & -     & Unimodal    & $256\times 256$ & US \\
DIAT~\cite{li2016deep}                    & arXiv 2016   & One-sided & Two-domain   & -     & Unimodal    & $128\times 128$ & TARR, CSIM, PSNR, SSIM \\
MPCR~\cite{lin2019image}                  & IJCAI 2019   & One-sided & Two-domain   & -     & Unimodal    & -               & FID, TARR, PSNR, US \\
CouncilGAN~\cite{nizan2020breaking}       & CVPR 2020    & One-sided & Two-domain   & -     & Multi-modal & $256\times 256$ & FID, KID \\
IcGAN~\cite{perarnau2016invertible}       & arXiv 2016   & One-sided & Multi-domain & Abs. Label & Unimodal    & $64\times 64$   & - \\
Fader Network~\cite{lample2017fader}      & NeurIPS 2017 & One-sided & Multi-domain & Abs. Label & Unimodal    & $256\times 256$ & US \\ 
UFDN~\cite{liu2018unified}                & NeurIPS 2017 & One-sided & Multi-domain & Abs. Label & Unimodal    & $64\times  64$  & TARR, L2 Error of Pixel Value, SSIM, PSNR \\
FPGAN~\cite{siddiquee2019learning}        & ECCV 2018    & Cyclic    & Multi-domain & Abs. Label & Unimodal    & $128\times 128$ & TARR, L1 Error of Pixel Value \\
SaGAN~\cite{zhang2018generative}          & ECCV 2018    & Cyclic    & Multi-domain & Abs. Label & Unimodal    & $128\times 128$ & ROC Curve for Face Verification \\
AttCycleGAN~\cite{lu2018attribute}        & ECCV 2018    & Cyclic    & Multi-domain & Abs. Label & Unimodal    & $128\times 128$ & SSIM \\
StarGAN~\cite{choi2018stargan}            & CVPR 2018    & Cyclic    & Multi-domain & Abs. Label & Unimodal    & $128\times 128$ & TARR, US \\
AttGAN~\cite{he2019attgan}                & TIP 2019     & One-sided & Multi-domain & Abs. Label & Unimodal    & $384\times 384$ & TARR, NAPR \\
ADSPM~\cite{wu2019attribute}              & ICCV 2019    & One-sided & Multi-domain & Abs. Label & Unimodal    & $512\times 512$ & FID, TARR, US \\
SDIT~\cite{wang2019sdit}                  & ACM MM 2019  & One-sided & Multi-domain & Abs. Label & Multi-modal & $128\times 128$ & LPIPS, TARR, CSIM \\
SMIT~\cite{romero2019smit}                & ICCVW 2019   & Cyclic    & Multi-domain & Abs. Label & Multi-modal & $128\times 128$ & IS, LPIPS, Conditional IS~\cite{huang2018multimodal} \\
HQGAN~\cite{deng2020controllable}         & TIFS 2020    & One-sided & Multi-domain & Abs. Label & Unimodal    & $512\times 512$ & FID, NAPR, MS-SSIM \\
INIT~\cite{cao2020informative}            & ECCV 2020    & One-sided & Multi-domain & Abs. Label & Unimodal    & $256\times 256$ & FID, TARR, US \\
StarGANv2~\cite{choi2020stargan}          & CVPR 2020    & Cyclic    & Multi-domain & Style Code & Multi-modal & $256\times 256$ & FID, LPIPS \\
HiSD~\cite{li2021image}                   & CVPR 2021    & Cyclic    & Multi-domain & Style Code & Multi-modal & $256\times 256$ & FID, US \\
RelGAN~\cite{wu2019relgan}                & ICCV 2019    & Cyclic    & Multi-domain & Rel. Label & Unimodal    & $256\times 256$ & FID, TARR, L1/L2 Error of Recon., SSIM, US \\
STGAN~\cite{liu2019stgan}                 & CVPR 2019    & One-sided & Multi-domain & Rel. Label & Unimodal    & $128\times 128$ & TARR, PSNR, SSIM, US \\
SSCGAN~\cite{chu2020sscgan}               & ECCV 2020    & One-sided & Multi-domain & Rel. Label & Unimodal    & $256\times 256$ & FID, TARR \\
CooGAN~\cite{chen2020coogan}              & ECCV 2020    & One-sided & Multi-domain & Rel. Label & Unimodal    & $768\times 768$ & TARR, PSNR, SSIM, US, Cost Analysis \\
GCN-reprs~\cite{bhattarai2020inducing}    & ECCV 2020    & One-sided & Multi-domain & Rel. Label & Unimodal    & $128\times 128$ & TARR, PSNR, SSIM \\
WarpGAN~\cite{dorta2020gan}               & CVPR 2020    & Cyclic    & Multi-domain & Rel. Label & Unimodal    & $128\times 128$ & TARR, CSIM, US \\
HifaGAN~\cite{gao2021high}                & CVPR 2021    & Cyclic    & Multi-domain & Rel. Label & Unimodal    & $256\times 256$ & FID, TARR, L1 Error of Pixel Value, QS~\cite{gu2020giqa} \\
\hline
\end{tabular}
}
\label{table:summary_image_domain_translation_methods}
\end{table*}

\subsubsection{Accuracy of Attribute Manipulation}

\textnormal{Target Attribute Recognition Rate} (TARR) is one of the most commonly used measurements to analyze whether the target attributes are retained in FAM results~\cite{perarnau2016invertible,liu2018unified,choi2018stargan,wu2019relgan,liu2019stgan,wang2019sdit,he2019attgan,deng2020controllable,lee2020maskgan,han2021disentangled,cho2019image,bhattarai2020inducing}.
For each target facial attribute $\alpha$ with desired value $l_{\alpha}^\ast$, an off-the-shelf attribute classifier $c_{\alpha}$ is usually adopted to estimate its label $l_{\alpha}$ in an edited image $\boldsymbol{x'}$, i.e., $l_{\alpha}=c_{\alpha}(\boldsymbol{x'})$.
Thus, TARR for the attribute $\alpha$ (denoted as $\text{TARR}_{\alpha}$) can be computed by
\begin{equation}
\text{TARR}_{\alpha} = \frac{1}{N} \sum_{\boldsymbol{x}} \boldsymbol{1}_{c_{\alpha}(\boldsymbol{x'})=l_{\alpha}^\ast}(\boldsymbol{x'})
\label{eq:TARR}
\end{equation}
where $N$ is the number of images in the test set, and $\boldsymbol{1}_{c_{\alpha}(\boldsymbol{x'})=l_{\alpha}^\ast}$ is an indicator function which returns 1 if the recognized attribute label equals to $l_{\alpha}^\ast$ otherwise 0.

The \textnormal{Non-Target Attribute Preservation Rate} (NAPR) is the counterpart of TARR, which measures whether non-target attributes are preserved in FAM results.
Following the notation of TARR, NAPR of a facial attribute we wish to preserve (denoted as $\bar{\alpha}$) is computed by
\begin{equation}
\text{NAPR}_{\bar{\alpha}} = \frac{1}{N} \sum_{\boldsymbol{x}} \boldsymbol{1}_{c_{\bar{\alpha}}(\boldsymbol{x'})=l_{\bar{\alpha}}^\ast}(\boldsymbol{x'})
\label{eq:NAPR}
\end{equation}

\subsubsection{Preservation of Image Semantics}

The {Error Caused by FAM} directly evaluates the deviation caused by applying FAM methods in various aspects, e.g., pixel values, coordinates of facial landmarks, and the angle of head pose. Commonly used quantitative measurements include the L1 norm, mean squared error (MSE), and root-mean-square error (RMSE).

The {Cosine Similarity} (CSIM) between identity embeddings extracted from input and output images measures the performance of preserving identity information. Given a pre-trained face recognition model $\mathcal{F}$ (e.g., ArcFace~\cite{deng2019arcface}), CSIM can be computed as
\begin{equation}
\text{CSIM} = cos(\mathcal{F}(\boldsymbol{x}), \mathcal{F}(\boldsymbol{x'}))
\label{eq:CSIM}
\end{equation}

The {Peak Signal-to-Noise Ratio} (PSNR)~\cite{wang2004image} is commonly used for measuring the image degradation in the reconstruction process (i.e., target attributes are the same as original ones), which can be computed as 
\begin{equation}
\text{PSNR} = 10\cdot \log_{10} \left(\frac{M^2}{E^2} \right)
\label{eq:PSNR}
\end{equation}
where $M$ denotes the maximum possible pixel value of an image and $E$ is the mean squared error between input and output images.

In contrast to the above-mentioned conventional metrics based on signal processing, the {Structural Similarity Index Measure} (SSIM)~\cite{wang2004image} measures the structural similarity between two images: 
\begin{equation}
\text{SSIM}(\boldsymbol{x},\boldsymbol{y}) = l(\boldsymbol{x},\boldsymbol{y})^{\alpha} \cdot c(\boldsymbol{x},\boldsymbol{y})^{\beta} \cdot s(\boldsymbol{x},\boldsymbol{y})^{\gamma}
\label{eq:SSIM}
\end{equation}
where $l(\boldsymbol{x},\boldsymbol{y})$, $c(\boldsymbol{x},\boldsymbol{y})$, and $s(\boldsymbol{x},\boldsymbol{y})$ are comparison measurements of luminance, contrast, and structure, respectively. $\alpha$, $\beta$, and $\gamma$ are the corresponding weights.
%
In addition, the (MS-SSIM)~\cite{wang2003multiscale} is a commonly used variant of SSIM where the structural similarity is computed over multiple scales to incorporate image details at different resolutions.

\subsubsection{Qualitative Measures}
In addition to quantitative measurements, a \textbf{User Study} (US) is also an important approach to evaluate the performance of FAM methods from the perspective of human perception~\cite{choi2018stargan,lee2018diverse,liu2019stgan,dorta2020gan,chang2020domain,choi2020stargan,tang2021attentiongan,han2021disentangled}.
Specifically, participants are asked to choose one from multiple FAM results obtained by different candidate methods which they consider suits a certain criterion the best.
The criteria can be flexibly designed to compare benchmark FAM methods in various dimensions.


\section{Image Translation}\label{sec:IDT_FAM}
\begin{figure}[t]
\begin{center}
\includegraphics[width=0.95\linewidth]{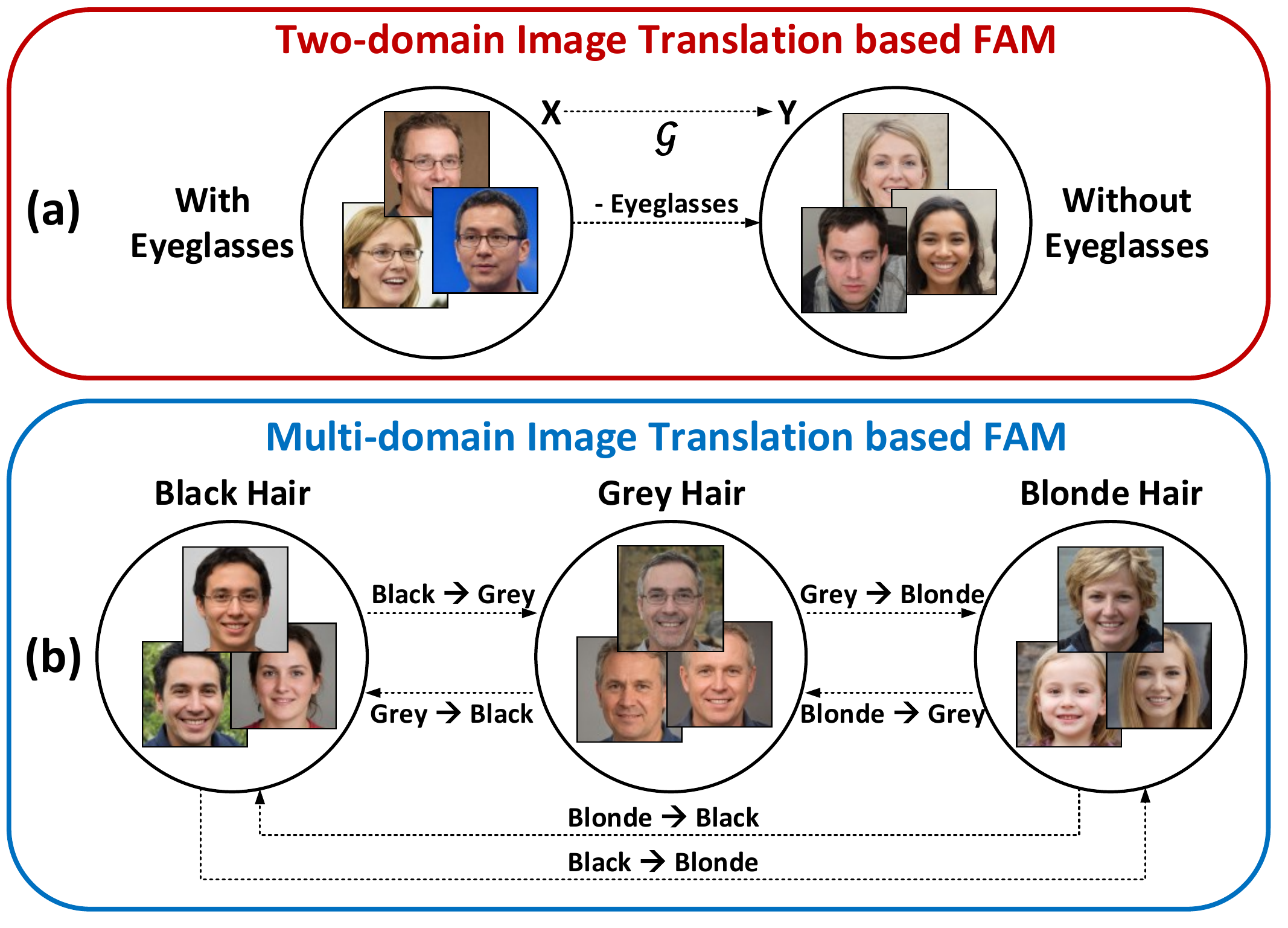}
\end{center}
\vspace{-0.2cm}
\caption{Illustration of (a) FAM between two domains, (b) FAM among $n$ domains ($n=3$, and six two-domain mappings have to be learned).
}
\vspace{-0.2cm}
\label{fig:domain_data_setting}
\end{figure}

Numerous methods consider FAM as an image-to-image translation (I2I) problem, where face images are grouped into visually distinct domains by their attributes.
According to the number of image domains involved, this class of methods can be generally divided into two-domain and multi-domains methods (see Fig.~\ref{fig:domain_data_setting}):
\begin{itemize}
  \item \textnormal{Two-domain Methods} can only manipulate one facial attribute between two image domains, and thus the model has to be repeatedly trained $n\cdot(n-1)$ times for manipulating $n$ facial attributes.
  \item \textnormal{Multi-domain Methods} estimate mapping functions across multiple domains (i.e., editing multiple attributes, each possible combination of the attributes specifies a unique domain) with one single framework, which greatly reduces the cost of training. 
\end{itemize}
A summary of image domain translation based FAM methods is provided in Table~\ref{table:summary_image_domain_translation_methods}.

\subsection{Two-domain Methods}

\begin{figure}[t]
\begin{center}
\includegraphics[width=1.0\linewidth]{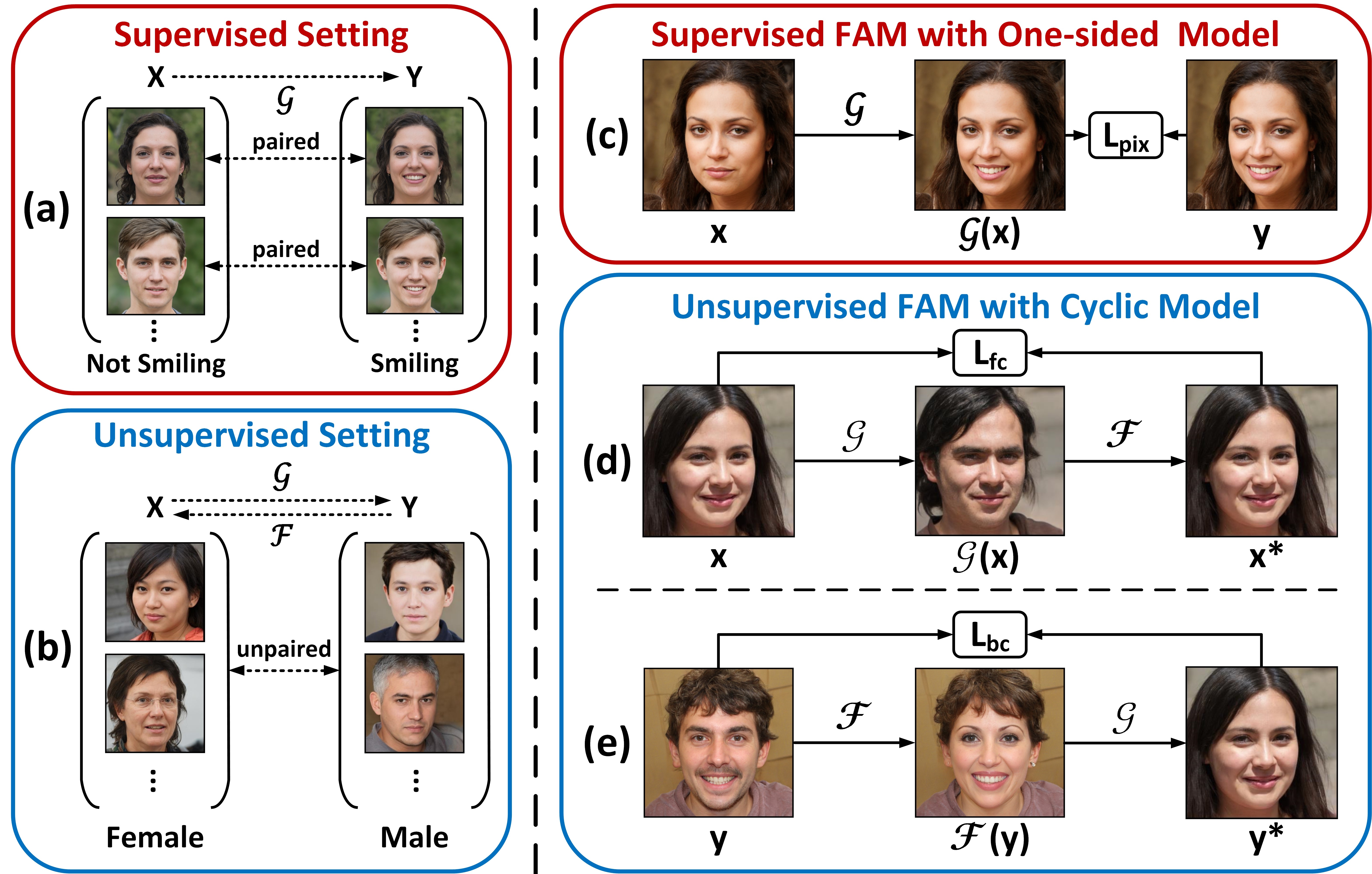}
\end{center}
\vspace{-0.2cm}
\caption{Illustration of (a) supervised and (b) unsupervised setting of training data in FAM, as well as the framework of (c) supervised FAM with the one-sided model trained on paired data, (d) forward translation process ($\boldsymbol{x}\rightarrow \mathcal{G}(\boldsymbol{x})\rightarrow \boldsymbol{x}^\ast$), and (e) backward translation process ($\boldsymbol{y}\rightarrow \mathcal{F}(\boldsymbol{y})\rightarrow \boldsymbol{y}^\ast$) in unsupervised FAM.
}
\vspace{-0.2cm}
\label{fig:pix2pix_vs_cycleGAN}
\end{figure}

The \textnormal{Pix2pix}~\cite{isola2017image} method uses an ordinary GAN-based model with the encoder-decoder architecture for image translation, i.e., $\mathcal{G}(\boldsymbol{x})=G(E(\boldsymbol{x}))$, and subsequent studies improve the resolution~\cite{wang2018high} and diversity~\cite{zhu2017toward} of translation results.
Although realistic translation results can be obtained by these methods, paired images $(\boldsymbol{x}, \boldsymbol{y})$ are required to compute the pixel-wise loss, i.e., $L_{pix}=\Vert \mathcal{G}(\boldsymbol{x}) - \boldsymbol{y}\Vert_2$, for supervising the training process, which are usually very expensive to collect for many facial attributes (e.g., gender, as shown in Fig.~\ref{fig:pix2pix_vs_cycleGAN}(b)).

FAM can be considered as an unsupervised image translation problem where the main objective is to regularize the learning process based on unaligned face images with no ground truth.
The \textnormal{cycle-consistency loss} proposed in \textnormal{CycleGAN}~\cite{zhu2017unpaired} (also in \textnormal{DiscoGAN}~\cite{kim2017learning}, \textnormal{DualGAN}~\cite{yi2017dualgan}, and \textnormal{UNIT}~\cite{liu2017unsupervised}) is one of the most widely used constraints in unsupervised I2I approaches. 
As shown in Fig.~\ref{fig:pix2pix_vs_cycleGAN}(d) and (e), $\mathcal{G}:X\rightarrow Y$ is coupled with an inverse mapping $\mathcal{F}:Y\rightarrow X$ to compute the cyclic reconstruction image $\boldsymbol{x}^\ast$, and a forward cycle loss $L_{fc}=\Vert \mathcal{F}(\mathcal{G}(\boldsymbol{x})) - \boldsymbol{x} \Vert_1$ is adopted to penalize the error between $\boldsymbol{x}$ and $\boldsymbol{x}^\ast$.
Similarly, a backward cycle loss can also be written as $L_{bc}=\Vert \mathcal{G}(\mathcal{F}(\boldsymbol{y})) - \boldsymbol{y} \Vert_1$, and the complete cycle-consistency constraint can be computed by summing up $L_{fc}$ and $L_{bc}$.

\begin{figure}[t]
\begin{center}
\includegraphics[width=0.95\linewidth]{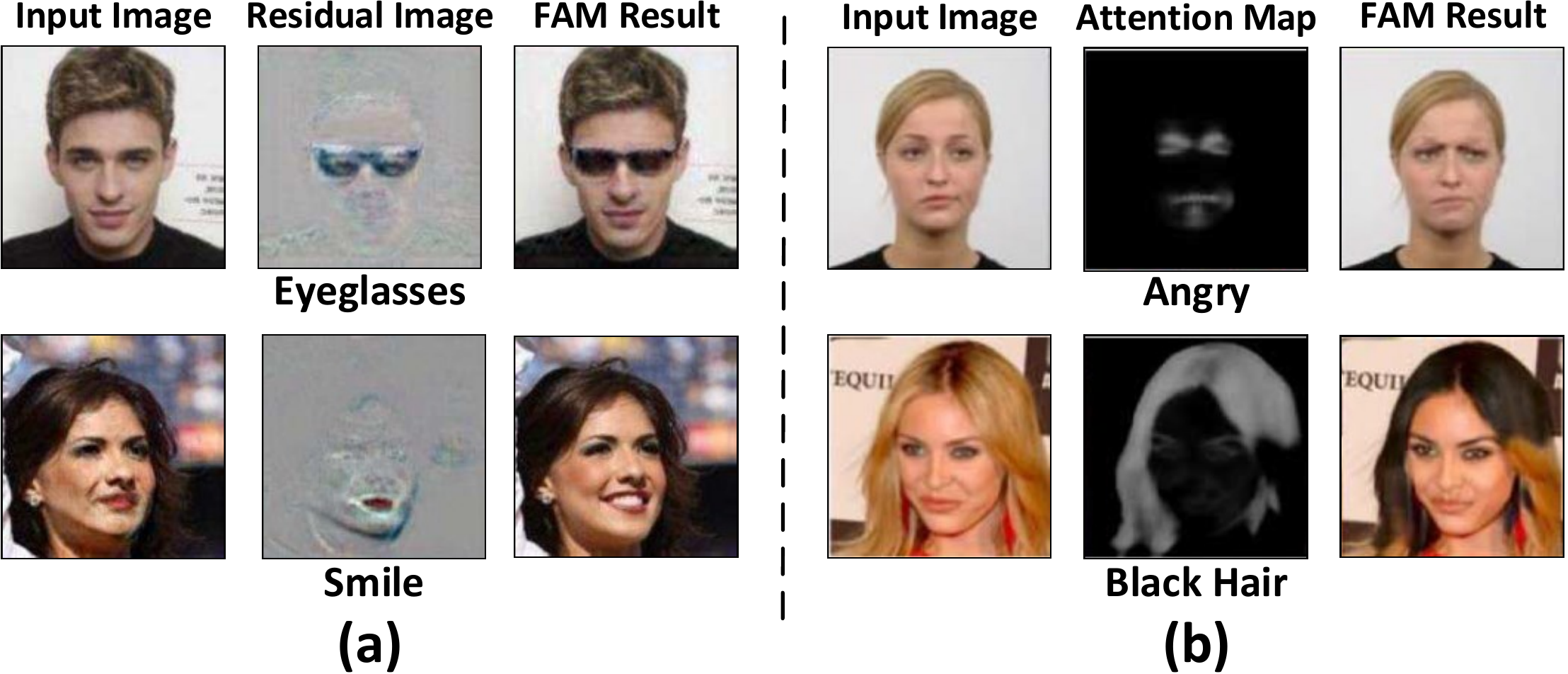}
\end{center}
\vspace{-0.2cm}
\caption{Sample results of a) ResGAN~\cite{shen2017learning} and b) AttentionGAN~\cite{tang2021attentiongan}, where darker region in the attention map indicates larger image modification. For each result, the target facial attribute is labeled underneath.}
\label{fig:ResGAN_AttentionGAN}
\vspace{-0.2cm}
\end{figure}

Numerous methods have been developed to improve CycleGAN and the cycle-consistency loss for I2I tasks.
Similar to~\cite{zhu2017toward}, \textnormal{AugCycleGAN}~\cite{almahairi2018augmented} proposes to incorporate auxiliary random variables for modeling data variation and generating diverse results.
To improves the spatial consistency between $\boldsymbol{x}$ and $\mathcal{G}(\boldsymbol{x})$, \textnormal{ResGAN}~\cite{shen2017learning} makes the translating network to only model the residual image, i.e., $\mathcal{G}(\boldsymbol{x})=\boldsymbol{x}+G(E(\boldsymbol{x}))$ (see Fig.~\ref{fig:ResGAN_AttentionGAN}(a)).
\textnormal{AttentionGAN}~\cite{tang2021attentiongan} further uses a spatial attention module to localize image changes, where a learnable attention mask $\boldsymbol{A}(\boldsymbol{x})\in [0,1]^{H\times W}$ ($H$ and $W$ denote the height and width of $\boldsymbol{x}$, respectively) is incorporated to denote the pixel-wise contribution of $\boldsymbol{x}$ to the FAM result (see Fig.~\ref{fig:ResGAN_AttentionGAN}(b)).
Thus, the translated result $\mathcal{G}(\boldsymbol{x})$ is computed by $\mathcal{G}(\boldsymbol{x})=\boldsymbol{A}(\boldsymbol{x})\cdot \boldsymbol{x} + (1-\boldsymbol{A}(\boldsymbol{x}))\cdot G(E(\boldsymbol{x}))$.
However, the pixel-level cycle-consistency constraint may not effectively deal with large geometric deformations in FAM (e.g., changing the hairstyle or removing eyeglasses).
To solve this problem, \textnormal{ACL-GAN}~\cite{zhao2020unpaired} introduces an additional discriminator to distinguish $\boldsymbol{x}$ between $\boldsymbol{x}^\ast$, which is adversarially trained with the translation network to enforce cycle-consistency from the data distribution perspective.

Other approaches aim to solve the unsupervised FAM problem with one-sided (i.e., non-cyclic) models, which reduces the cost of training a reverse mapping function.
To this end, additional loss functions are usually incorporated to regulate the under-constrained mapping $\mathcal{G}$.
Aside from the pixel-wise attention mechanism in the translation network, \textnormal{DIAT}~\cite{li2016deep} enforces the high-level consistency between $\boldsymbol{x}$ and $\mathcal{G}(\boldsymbol{x})$ with the perceptual loss and identity loss.
\textnormal{MPCR}~\cite{lin2019image} introduces an intermediate image domain $X'$, and enforces the consistency between the translation results through two paths, i.e., $X\rightarrow Y$ and $X\rightarrow X'\rightarrow Y$.
In addition, \textnormal{CouncilGAN}~\cite{nizan2020breaking} models the mapping $\mathcal{G}$ with multiple translation networks and discriminators, which enables generating diverse FAM results.

\subsection{Multi-domain Methods}\label{sec:multi_domain}

To achieve multi-domain FAM with a single framework, one straightforward solution is to aggregate multiple two-domain networks into a single model~\cite{hui2018unsupervised,zhang2018sparsely,zhao2018modular}.
However, computational loads would heavily increase as more facial attributes are considered.
Thus, numerous approaches perform multi-domain FAM with a single pair of encoder and generator, where target attributes are indicated using label vectors.

The label vector, denoted as $\boldsymbol{l}=[l_1, l_2, \ldots, l_N]\in \mathbb{R}^N$, where $l_i$ $(i=1,2,\ldots,N)$ specifies the state of the $i$-th attribute, is one of the most widely used representations of facial attributes in the multi-domain FAM methods due to its flexibility and interpretability.
According to the definition of $l_i$, label vectors can be further divided into two types, i.e., \textnormal{absolute label vectors} and \textnormal{relative label vectors}.

\subsubsection{Absolute Label Vectors}\label{sec:abs_label_vec}
\begin{figure}[t]
\begin{center}
\includegraphics[width=1.0\linewidth]{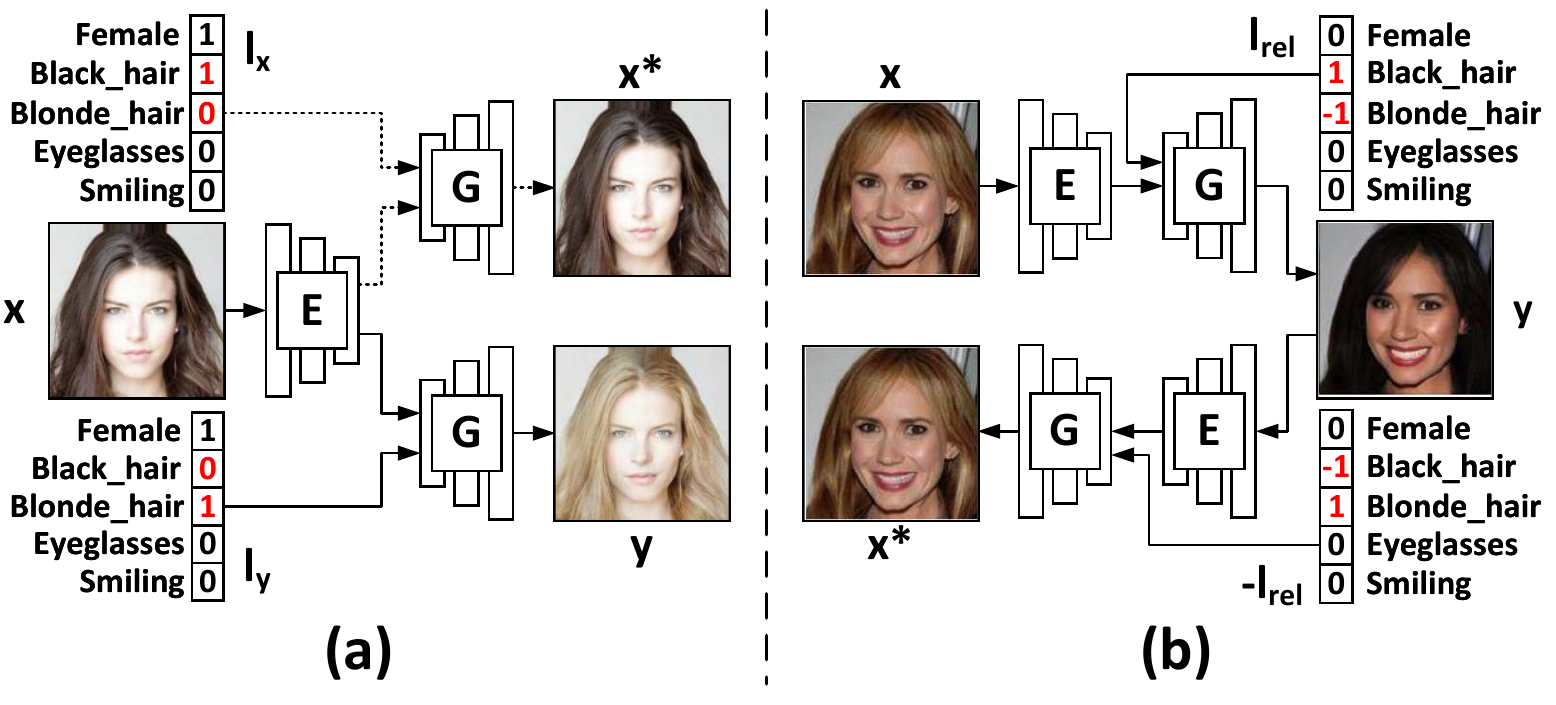}
\end{center}
\vspace{-0.2cm}
\caption{Illustration of FAM conditioned on (a)  absolute and (b) relative attribute vectors. 
(a) $\boldsymbol{x^\ast}$ denotes the result of identity mapping and $\boldsymbol{y}$ the result of FAM. 
(b) $\boldsymbol{l_{rel}}$ indicates the attributes to be edited, and $-\boldsymbol{l_{rel}}$ can be naturally used to guide the reconstruction process.
}
\label{fig:abs_rel_compare}
\vspace{-0.2cm}
\end{figure}

Absolute label vectors describe the actual state of attributes for face images.
As one of the most earliest attempts, \textnormal{IcGAN}~\cite{perarnau2016invertible} uses label vectors with binary values as the conditional information, i.e, $\boldsymbol{l}=\{0,1\}^N$, where $l_i=1$ ($l_i=0$) indicates the presence (absence) of the $i$-th facial attribute (see Fig.~\ref{fig:abs_rel_compare}(a)).
Specifically, one encoder network $E$  embeds the input image $\boldsymbol{x}$ from any domain into the latent space, and one decoding network $G$ generates the FAM result according to the target attribute vector $\boldsymbol{l_y}$, i.e., $\boldsymbol{y}=G(E(\boldsymbol{x}), \boldsymbol{l_y})$.
To improve the consistency between $\boldsymbol{x}$ and the FAM results, \textnormal{AttGAN}~\cite{he2019attgan} introduces an identity mapping loss $L_{idt}$ to better preserve the image content in $\boldsymbol{x}$ irrelevant to the target attributes:
\begin{equation}
L_{idt}=\Vert G(E(\boldsymbol{x}), \boldsymbol{l}_x)-\boldsymbol{x} \Vert_1 \label{eq:idt_loss}
\end{equation}
where $G(E(\boldsymbol{x}), \boldsymbol{l}_x)$ refers to the identity mapping result of $\boldsymbol{x}$ conditioned on the label vector of original attributes $\boldsymbol{l}_x$ (denoted as $\boldsymbol{x}^\ast$ in Fig.~\ref{fig:abs_rel_compare}(a)).
Moreover, making use of a classification network $C$, a cross-entropy based attribute classification loss $L_{cls}$ is also adopted to ensure the target attribute is modified properly:
\begin{equation}
L_{cls}=\sum^n_{i=1}-\boldsymbol{l}_y^{(i)}\log C^{(i)}(\boldsymbol{y}) - (1-\boldsymbol{l}_y^{(i)})\log (1-C^{(i)}(\boldsymbol{y}))
\end{equation}
Using AttGAN as the backbone, \textnormal{HQGAN}~\cite{deng2020controllable} achieves FAM on high-resolution face images ($512\times 512$) with the aid of a wavelet-based perceptual loss for improving the high-frequency textural details in translation results.

In \textnormal{StarGAN}~\cite{choi2018stargan},
a cycle-consistency loss $L_{cyc}$ conditioned on attribute labels is exploited: 
\begin{equation}
L_{cyc}=\Vert G(E(\boldsymbol{y}), \boldsymbol{l_x})-\boldsymbol{x} \Vert_1
\end{equation}
where $G(E(\boldsymbol{y}), \boldsymbol{l_x})$ is the cycle-reconstruction of $\boldsymbol{x}$.
An attribute classification loss $L_{cls}$ is also used in StarGAN, which is implemented by multi-tasking the discriminator $D$ instead of using an additional network as in AttGAN.
Moreover, StarGAN augments the label vector with masks, which facilitates training and inference processes with multiple datasets containing different attribute labels.
\textnormal{FPGAN}~\cite{siddiquee2019learning} improves StarGAN by introducing the identity mapping loss in Eq.~\ref{eq:idt_loss} to the translation in both directions, which helps reduce unnecessary image modifications.
In addition, \textnormal{SaGAN}~\cite{zhang2018generative} introduces the spatial attention mechanism to localize image changes as in~\cite{tang2021attentiongan}, where the attention map $\boldsymbol{A}(\boldsymbol{x})$ is dependent on the target attribute label $\boldsymbol{l_y}$.
On the other hand, \textnormal{AttCycleGAN}~\cite{lu2018attribute} adapts StarGAN to a face image super-resolution task where label vectors are used to control the attribute of generated HR images.

The disentanglement between the latent embedding of $\boldsymbol{x}$ (i.e., $\boldsymbol{z}=E(\boldsymbol{x})$) and the semantic information contained in $\boldsymbol{l_y}$ has also been studied.
Both \textnormal{Fader Network}~\cite{lample2017fader} and \textnormal{UFDN}~\cite{liu2018unified} use an auxiliary latent discriminator $D_Z$ to identify the true label of $\boldsymbol{x}$ given $\boldsymbol{z}$ while the encoder $E$ is trained to confuse $D_Z$.
This adversarial learning process ensures that $\boldsymbol{z}$ is attribute-invariant, and thus the semantic of FAM results (i.e., $\boldsymbol{y}=G(\boldsymbol{z}, \boldsymbol{l_y})$) is guaranteed to be determined only by the information encoded in $\boldsymbol{l_y}$ and the influence of attributes in $\boldsymbol{x}$ is minimized.

To improve the accuracy of facial attribute translations with large geometric deformation, \textnormal{ADSPM}~\cite{wu2019attribute} proposes a spontaneous motion estimation module to model the motion field for rendering attribute-driven shape transformation, e.g., change of expressions or shape of facial components, via an explicit warping operation.
To deal with large changes in facial geometry, \textnormal{INIT}~\cite{cao2020informative} proposes to improve the visual quality of FAM results with large domain variations. 
Specifically, an adversarial importance weighting technique is designed to select more informative samples for training, and a multi-hop sample training strategy is used to stabilize the training process.

\begin{table*}[t]
\caption{Summary of the characteristics of facial semantic decomposition based FAM Methods. Disentanglement indicates how the disentanglement of different facial semantics is achieved.}
\vspace{-0.2cm}
\rowcolors{1}{}{lightgray}
\centering
\ra{1.3}
\resizebox{2.0\columnwidth}{!}{%
\begin{tabular}{cccccc}
\hline
Method Name                      & Publication  & Decomposition Level   & Disentanglement & Resolution & Quantitative Metrics \\
\hline
MUNIT~\cite{huang2018multimodal} & ECCV 2018    & Binary-domain  & Cross Cycle-Consistency Loss & $256\times 256$ & Conditional IS, LPIPS, US \\
cd-GAN~\cite{lin2018conditional} & CVPR 2018    & Binary-domain  & Cross Cycle-Consistency Loss & $64\times 64$   & US \\
GDWCT~\cite{cho2019image}        & CVPR 2019    & Binary-domain  & Cross Cycle-Consistency Loss & $256\times 256$ & TARR, US, Inference Time \\ 
DRIT~\cite{lee2018diverse}       & ECCV 2018    & Binary-domain  & Adversarial Training         & $256\times 256$ & LPIPS, TARR, L1 Error of Self Recon., US \\
DMIT~\cite{yu2019multi}          & NeurIPS 2019 & Multi-domain   & Adversarial Training         & $256\times 256$ & FID, LPIPS, PSNR, SSIM, US \\
DRIT++~\cite{lee2020drit++}      & IJCV 2020    & Multi-domain   & Adversarial Training         & $216\times 216$ & FID, LPIPS, US, L2 Distance of Latent Embeddings \\
DosGAN~\cite{lin2019exploring}     & TIP 2019  & Multi-domain    & Pre-trained Classifier       & $256\times 256$ & FID, TARR, CSIM, PSNR, SSIM \\
D$\mathbf{^2}$AE~\cite{liu2018exploring}  & CVPR 2018 & Binary-semantic & Adversarial Training  & $235\times 235$ & TARR, ROC Curve of Face Verification \\
LORD~\cite{gabbay2019demystifying} & ICLR 2019 & Binary-semantic & Latent Optimization & $128\times 128$   & LPIPS, TARR \\
OverLORD~\cite{gabbay2021scaling}  & ICCV 2021 & Binary-semantic & Latent Optimization & $1024\times 1024$ & FID, LPIPS, TARR, L2 Error of Pose, CSIM \\
LSM~\cite{nitzan2020face}          & TOG 2020  & Binary-semantic & Latent Space of Pre-trained StyleGAN & $1024\times 1024$ & FID, CSIM, L2 Error of Landmarks and Pose, PSNR \\
SwapAE~\cite{park2020swapping}           & NeurIPS 2020 & Binary-semantic & Style Coherence Supervision          & $1024\times1024$   & LPIPS, US, Time Cost, SIFID-SSD Curve~\cite{kolkin2019style,shaham2019singan} \\
Sensorium~\cite{nederhood2021harnessing} & ICCV 2021    & Binary-semantic & Pre-defined Content Representation   & $256\times256$     & FID \\
SofGAN~\cite{chen2022sofgan}             & TOG 2022     & Binary-semantic & Pre-defined Geometric Representation & $1024\times1024$   & FID, LPIPS, CSIM \\
SNI~\cite{alharbi2020disentangled}       & CVPR 2020    & Binary-semantic & Decomposed Latent Space              & $1024\times1024$   & FID, LPIPS, linear separability in $Z$ and $W$ \\
DAT~\cite{kwon2021diagonal}              & ICCV 2021    & Binary-semantic & Decomposed Latent Space              & $1024\times1024$   & FID, LPIPS, PPL~\cite{karras2019style} \\
TransEditor~\cite{xu2022transeditor}     & CVPR 2022    & Binary-semantic & Decomposed Latent Space    & $256\times256$     & LPIPS, Attribute Re-scoring Analysis \\
DNA-GAN~\cite{xiao2018dna}               & ICLRW 2018   & Multi-semantic  & Contrastive Training Batch & $128\times128$     & - \\
ELEGANT~\cite{xiao2018elegant}           & ECCV 2018    & Multi-semantic  & Contrastive Training Batch & $256\times 256$    & FID \\
GAN-Control~\cite{shoshan2021gan}        & ICCV 2021    & Multi-semantic  & Contrastive Training Batch & $512\times 512$    & FID, TARR, CSIM, US \\
DiscoFaceGAN~\cite{deng2020disentangled} & CVPR 2020    & Multi-semantic  & 3D Graphics Model           & $256\times 256$    & FID, PPL~\cite{karras2019style}, Disentanglement Score \\
GIF~\cite{ghosh2020gif}                  & 3DV 2020     & Multi-semantic  & 3D Graphics Model           & $1024\times 1024$  & FID, L2 Error of Face Mesh Vertices, US \\
VariTex~\cite{buhler2021varitex}         & ICCV 2021    & Multi-semantic  & 3D Graphics Model           & $256\times 256$    & FID, CSIM, US \\
ConfigNet~\cite{kowalski2020config}      & ECCV 2020    & Multi-semantic  & 3D Graphics Model           & $256\times 256$    & FID, TARR, NAPR \\
NFENet~\cite{shu2017neural}              & CVPR 2017    & Multi-semantic  & 3D Graphics Model           & $64\times 64$      & - \\
MOST-GAN~\cite{medin2022most}            & AAAI 2022    & Multi-semantic  & 3D Graphics Model           & $256\times 256$    & Mean Absolute Error of Head Pose \\
MaskGuidedGAN~\cite{gu2019mask}          & CVPR 2019    & Multi-semantic  & Face Parsing Map           & $256\times 256$    & FID, Face Parsing Accuracy \\
MaskGAN~\cite{lee2020maskgan}            & CVPR 2020    & Multi-semantic  & Face Parsing Map           & $512\times 512$    & FID, TARR, CSIM, US, Face Parsing Accuracy \\
SemanticStyleGAN~\cite{shi2022semanticstylegan} & CVPR 2022 & Multi-semantic  & Face Parsing Map       & $512\times 512$    & IS, FID, LPIPS, L1 Error of Pixel Value \\
InfoGAN~\cite{chen2016infogan}           & NeurIPS 2016 & Multi-semantic  & Unsupervised Factorization & $64\times 64$      & Lower Bound of Mutual Information \\
$\beta$-VAE~\cite{higgins2017beta}       & ICLR 2017    & Multi-semantic  & Unsupervised Factorization & $64\times 64$      & Disentanglement Metric Score \\
FactorVAE~\cite{kim2018disentangling}    & ICML 2018    & Multi-semantic  & Unsupervised Factorization & $64\times 64$      & L2 Error of Recon., Disentanglement Metrics, Total Correlation \\
HoloGAN~\cite{nguyen2019hologan}         & ICCV 2019    & Multi-semantic  & Unsupervised Factorization & $128\times 128$    & KID \\
SCM~\cite{chen2019semantic}              & CVPR 2019    & Multi-semantic  & Unsupervised Factorization & $448\times 448$    & US, Inference Time \\
EIS~\cite{collins2020editing}            & CVPR 2020    & Multi-semantic  & Unsupervised Factorization & $1024\times 1024$  & FID, L2 Error of Pixel Value \\
StyleFusion~\cite{kafri2022stylefusion}  & TOG 2022     & Multi-semantic  & Unsupervised Factorization & $1024\times 1024$  & FID \\
RIS~\cite{chong2021retrieve}             & ICCV 2021    & Multi-semantic  & Unsupervised Factorization & $1024\times 1024$  & FID, Attribute Matching Score, TRSI-IoU \\
\hline
\end{tabular}
}
\label{table:summary_semantic_decomposition_methods}
\end{table*}

\begin{figure}[t]
\begin{center}
\includegraphics[width=1.0\linewidth]{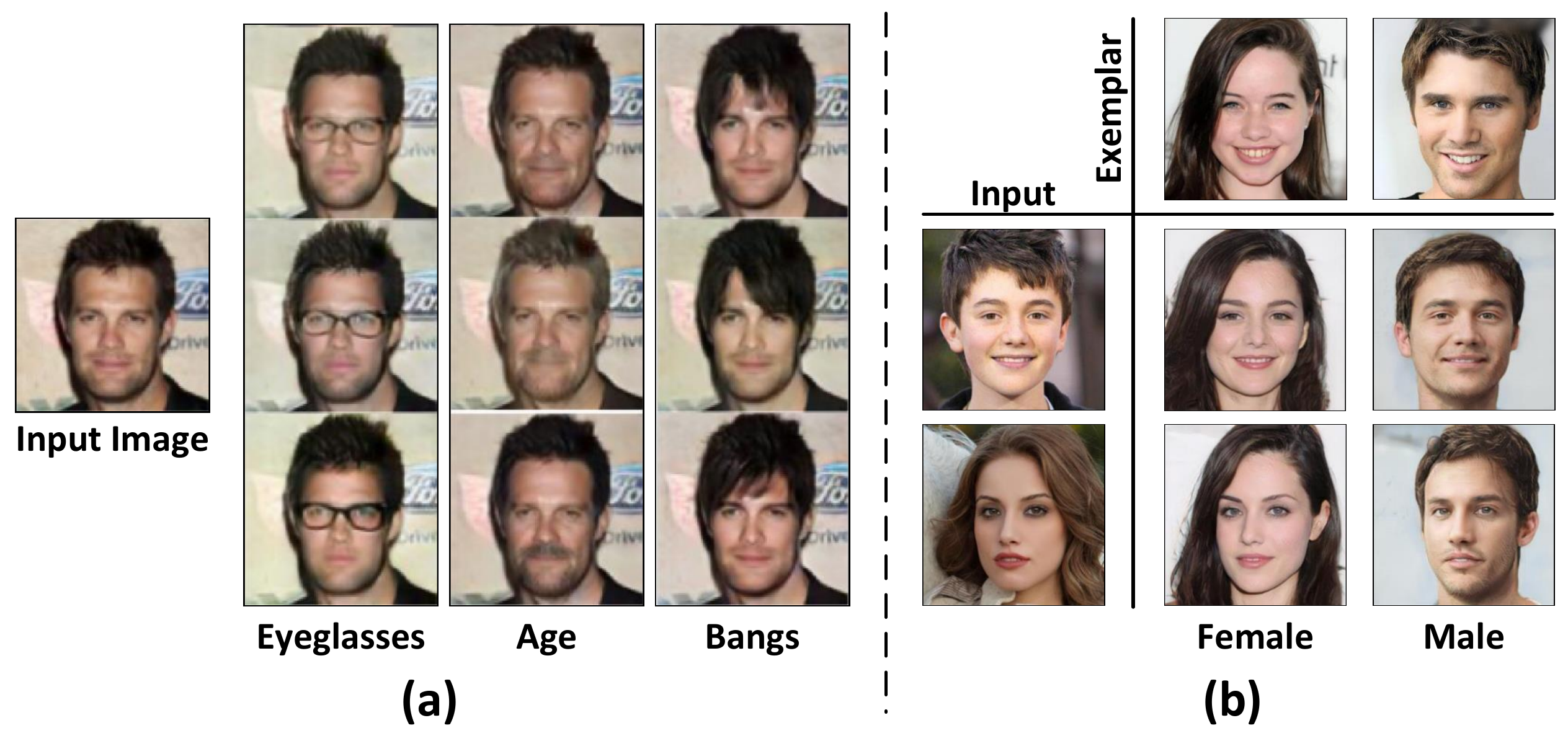}
\end{center}
\vspace{-0.2cm}
\caption{Sample multi-modal FAM results by (a) SDIT~\cite{wang2019sdit} with latent code re-sampling and (b) StarGANv2~\cite{choi2020stargan} guided by exemplars.}
\vspace{-0.2cm}
\label{fig:SDIT_StarGANv2}
\end{figure}

For diverse translation results, \textnormal{SDIT}~\cite{wang2019sdit} integrates a random latent code $\boldsymbol{z}$ sampled from a Gaussian prior to into the generator, such that re-sampling $\boldsymbol{z}$ at test time would produce multi-modal FAM results (as shown in Fig.~\ref{fig:SDIT_StarGANv2}(b)).
Similarly, \textnormal{SMIT}~\cite{romero2019smit} proposes a domain embedding module to compute the guiding signal for FAM based on the target label vector $\boldsymbol{l_y}$ and a random style code $\boldsymbol{z}\sim \mathcal{N}(0,1)$.
To further improve the controllability of multi-modal FAM results, an encoder network $E_s$ is incorporated in \textnormal{StarGANv2}~\cite{choi2020stargan} for computing the attribute-specific style representation of exemplar images, which is used to guide the generation of FAM results.
Although diverse FAM results with controlled semantics can be obtained, irrelevant facial attributes (e.g., age and hair color) have also been modified, as shown in Fig.~\ref{fig:SDIT_StarGANv2}(b).
Thus, to improve the disentanglement between different facial attributes, \textnormal{HiSD}~\cite{li2021image} improves StarGANv2 by organizing facial attributes into a hierarchical structure with disentangled style allocation, and each attribute is associated with a dedicated translator network which produces diverse and localized modifications.

\subsubsection{Relative Label Vectors}\label{sec:rela_label_vec}

Several FAM methods use label vectors to indicate the \textnormal{difference} in facial attributes between input and desired output images, i.e., relative label vectors.
\textnormal{RelGAN}~\cite{wu2019relgan} shows that the main problem of absolute label vector is that it requires specifying the entire set of attributes, which is inefficient since in practice most of them remain unchanged during FAM.
To address this problem, RelGAN exploits the relative label vector, i.e., $\boldsymbol{l_{rel}}=\boldsymbol{l_y}-\boldsymbol{l_x}$, to inform the model what attributes to modify, and $-\boldsymbol{l_{rel}}$ naturally describes the reverse mapping process (see Fig.~\ref{fig:abs_rel_compare}(b)).
To ensure that the attribute difference between $\boldsymbol{x}$ and the FAM result $\boldsymbol{y}$ matches $\boldsymbol{l_{rel}}$, instead of using the classification loss $L_{cls}$ as in AttGAN or StarGAN, an auxiliary discriminator is adversarially trained to distinguish between real triplets $(\boldsymbol{x}, \boldsymbol{l_{rel}}, \boldsymbol{x}')$ ($\boldsymbol{x}'$ is a real image with the target attribute) and fake ones $(\boldsymbol{x}, \boldsymbol{l_{rel}}, \boldsymbol{y})$.

The relative label vector is also used in other FAM studies as conditional information.
To help preserve the irrelevant content in input images without weakening the ability of attribute translation, \textnormal{SSCGAN}~\cite{chu2020sscgan} uses skip connections to transfer style features and spatial information in input images.
In addition, skip connections are also incorporated in \textnormal{STGAN}~\cite{liu2019stgan} and \textnormal{HifaGAN}~\cite{gao2021high} to facilitate the forwarding of textural information in input images to FAM results, which are implemented with selective transfer units and wavelet-transformation blocks.
In~\cite{chen2020coogan}, \textnormal{CooGAN} proposes a dual-path framework to perform FAM on HR images, where one branch focuses on generating fine-grained facial patches and the other monitors the overall facial structure.
On the other hand, \textnormal{GCN-reprs}~\cite{bhattarai2020inducing} uses Graph Convolutional Networks to extract more structured information of high-level semantics and the relative dependencies of facial attributes, which is used to guide both the generator and discriminator for enhanced FAM results.
In~\cite{dorta2020gan}, \textnormal{WarpGAN} further extends ADSPM~\cite{wu2019attribute} to manipulate HR face images with relative attributes, which is claimed to be beneficial for generating localized image edits.

\section{Semantic Decomposition} \label{sec:FSD_FAM}

\begin{figure}[t]
\begin{center}
\includegraphics[width=0.85\linewidth]{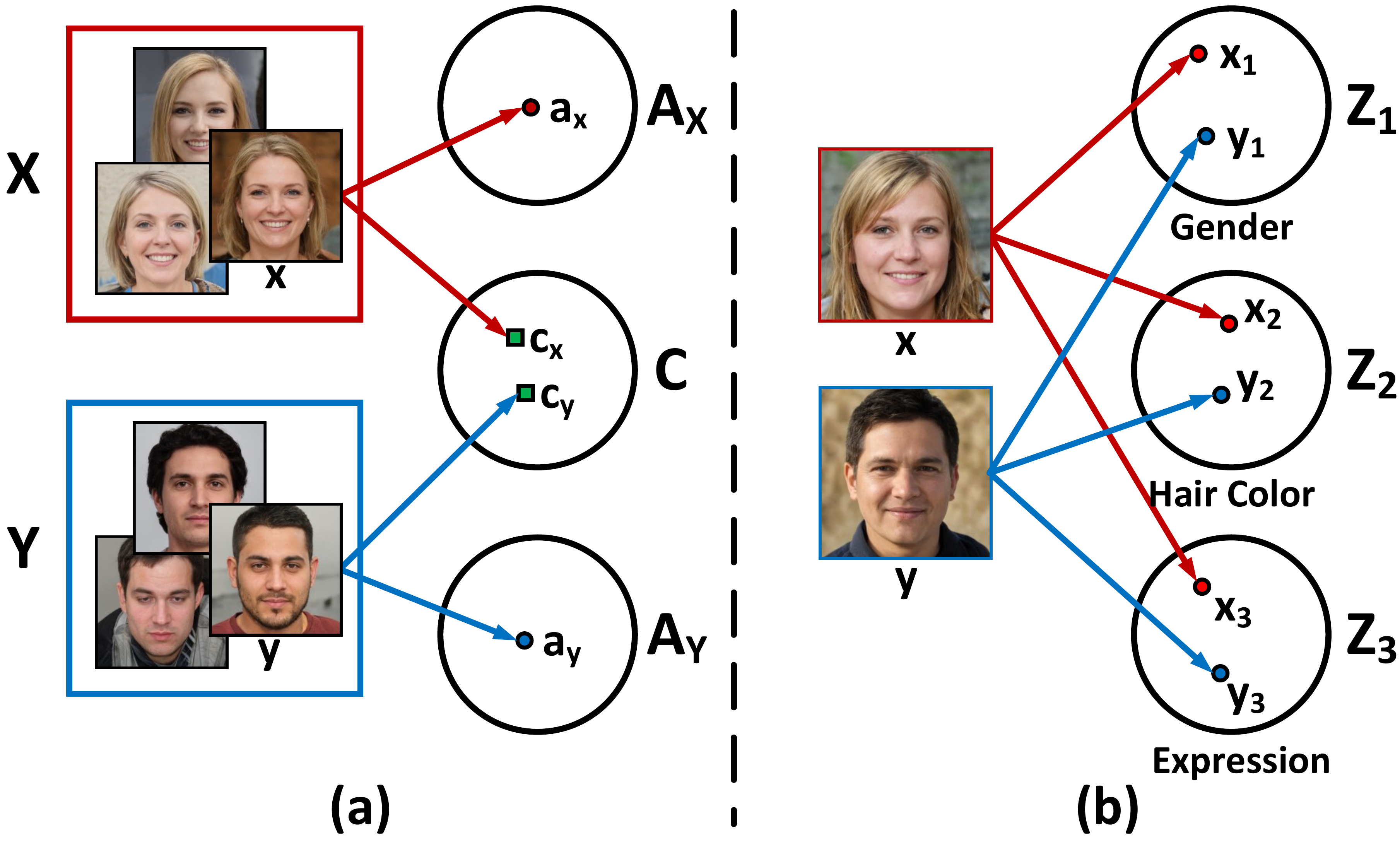}
\end{center}
\vspace{-0.2cm}
\caption{Illustration of (a) domain-level latent decomposition ($X$: smiling female subjects with blonde hair, $Y$: male subjects with black hair and neutral expression), and (b) instance-level latent decomposition with three facial semantics considered.
}
\label{fig:domain_vs_instance}
\vspace{-0.2cm}
\end{figure}

Semantic decomposition approaches encode the input image $\boldsymbol{x}$ into multiple separate latent spaces, where the embedded representations are used to control different facial features.
FAM can be performed by editing the latent code associated with target attributes.
%
Based on the granularity of problem formulation, facial semantic decomposition can be further divided into two categories, i.e., \textnormal{domain-level decomposition} and \textnormal{instance-level decomposition}:
\begin{itemize}
  \item \textnormal{Domain-level decomposition} assumes that each image, regarded as a domain member, can be encoded into two separate latent spaces, one for modeling the \textnormal{domain-invariant} information (i.e., \textnormal{Content Space} $C$), and the other for capturing the \textnormal{domain-specific} variations (i.e., \textnormal{Attribute Space} $A$), as shown in Fig.~\ref{fig:domain_vs_instance}(a).
  \item \textnormal{Instance-level decomposition} focus on factorizing  facial semantics of individual input images into separate latent components (e.g., $Z_1$, $Z_2$, and $Z_3$ in Fig.~\ref{fig:domain_vs_instance}(b)). %
  This enables more flexible and disentangled control of facial attributes in finer granularity compared to learning domain-level translation patterns.
\end{itemize}
A summary of facial semantic decomposition based FAM methods is provided in Table~\ref{table:summary_semantic_decomposition_methods}.

\subsection{Domain-level Decomposition}\label{sec:domain_decomp}
\begin{figure}[t]
\begin{center}
\includegraphics[width=0.90\linewidth]{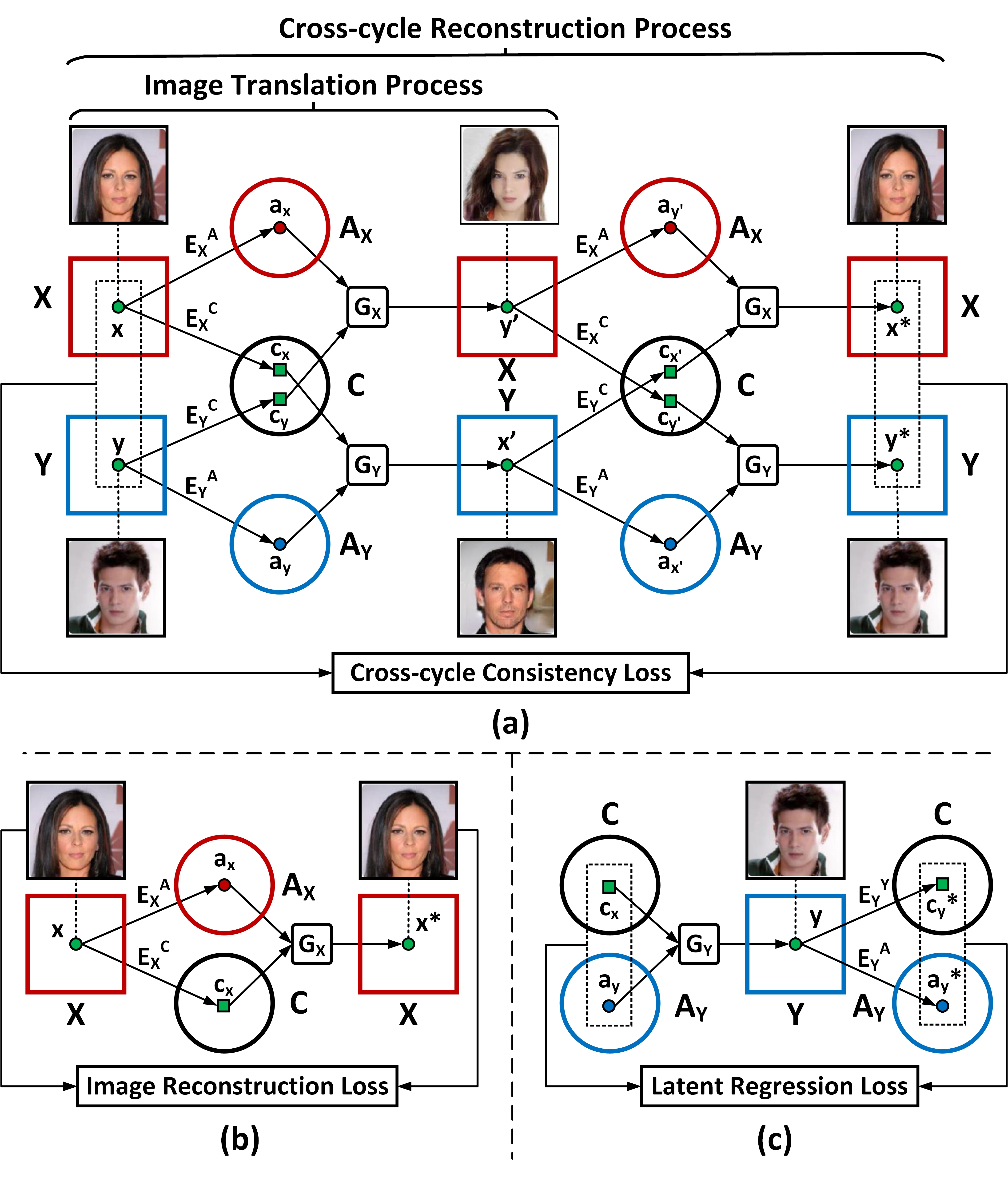}
\end{center}
\vspace{-0.2cm}
\caption{Overview of the (a) image translation ($\boldsymbol{x}\rightarrow \boldsymbol{x}'$, $\boldsymbol{y}\rightarrow \boldsymbol{y}'$) and cross-cycle reconstruction process (forward: $\boldsymbol{x}\rightarrow \boldsymbol{x}'\rightarrow \boldsymbol{x}$, backward: $\boldsymbol{y}\rightarrow \boldsymbol{y}'\rightarrow \boldsymbol{y}$) in manipulating the Gender attribute ($X$: Female with \textnormal{{\color{red} red}} boxes, and $Y$: Male with \textnormal{{\color{blue} blue}} boxes), (b) image reconstruction process, and (c) latent regression process. Note that the image reconstruction loss and latent regression loss are computed for both domains, and we only show their applications on one domain for conciseness. Images are directly obtained from~\cite{cho2019image} for illustrative purposes.
}
\vspace{-0.2cm}
\label{fig:decomposed_latent_models}
\end{figure}

Based on the content space $C$ and attribute space $A$, the latent embeddings of an image $\boldsymbol{x}\in X$, denoted as $\boldsymbol{c_x}$ and $\boldsymbol{a_x}$, represent the semantic component that should be preserved and to be manipulated.
As shown in Fig.~\ref{fig:decomposed_latent_models}(a), given an exemplar image $\boldsymbol{y}$ in the target domain $Y$, the translation of $\boldsymbol{x}$ (denoted as $\boldsymbol{x}'$) can be obtained based on its content embedding $\boldsymbol{c_x}$ and the attribute code of $\boldsymbol{y}$ (i.e., $\boldsymbol{a_y}$), and $\boldsymbol{y}$ can be similarly translated into $\boldsymbol{y}'\in X$.

To learn a content-attribute decomposition between $C$ and $A_X$/$A_Y$, \textnormal{MUNIT}~\cite{huang2018multimodal} uses an image reconstruction loss to ensure that all information in $\boldsymbol{x}$ is contained in $\boldsymbol{c_x}$ and $\boldsymbol{a_x}$ (see Fig.~\ref{fig:decomposed_latent_models}(b),
similar for $\boldsymbol{y}$).
Moreover, a latent regression loss is  incorporated to guarantee that encoders and generators are inverses of each other (Fig.~\ref{fig:decomposed_latent_models}(c)), which also implicitly facilitates t disentanglement between $\boldsymbol{c_x}$ and $\boldsymbol{a_x}$.
On the other hand, \textnormal{DRIT}~\cite{lee2018diverse} adopts an auxiliary content discriminator $D_C$ to distinguish the domain membership of $\boldsymbol{c_x}$ and $\boldsymbol{c_y}$, and the encoder networks are trained to confuse $D_C$ by making $\boldsymbol{c_x}$ and $\boldsymbol{c_y}$ only contain the domain-variant information.
Leveraging such disentanglement, two cross-cycle consistency losses (Fig.~\ref{fig:decomposed_latent_models}(a)). are proposed to regulate the learned mappings between $X$ and $Y$:
\begin{align}
&L_{CC}^{X\rightarrow Y\rightarrow X} = \Vert G_X(\boldsymbol{c_{x'}}, \boldsymbol{s_{y'}})) - \boldsymbol{x}\Vert_1 \label{eq:decomp_latent_cross_recon_xyx}\\
&L_{CC}^{Y\rightarrow X\rightarrow Y} = \Vert G_Y(\boldsymbol{c_{y'}}, \boldsymbol{s_{x'}})) - \boldsymbol{y}\Vert_1 \label{eq:decomp_latent_cross_recon_yxy}
\end{align}
These constraints are also proposed in \textnormal{cd-GAN}~\cite{lin2018conditional} from the perspective of dual learning.
In~\cite{cho2019image}, a model similar to DRIT is developed where a group-wise whitening-and-coloring transformation is exploited to combine the embedding of content and attributes, which improves the memory and time efficiency as well as the quality of generation results.

For the multi-domain setting, numerous approaches model images from different domains with a common auto-encoder network, which is similar to the methods discussed in Section~\ref{sec:multi_domain}.
\textnormal{DRIT++}~\cite{lee2020drit++} extends DRIT~\cite{lee2018diverse} to receive one-hot domain code at the input of attribute encoders and generators for specifying the source and target domain, and the discriminator is designed to classify the domain of given images in addition to distinguishing the realism.
In~\cite{yu2019multi}, \textnormal{DMIT} uses an additional encoder network parallel to the content and attribute encoders for computing the domain label of input images, and the generator is extended to also receive such domain labels as input to control the domain membership of output images.
On the other hand, \textnormal{DosGAN}~\cite{lin2019exploring} adopts a single pre-trained domain-classifier to extract the attribute feature specific to each image domain, which facilitates disentangling embeddings in the content and attribute space.

\subsection{Instance-level Decomposition}\label{sec:inst_decomp}

%
Instance-level FAM methods can be divided into \textnormal{binary semantic decomposition} and \textnormal{multiple semantic decomposition} according to the number of facial attributes considered.

\subsubsection{Binary Semantic Decomposition}\label{sec:bi_seman_decomp}

FAM methods based on binary semantic decomposition represent an image with two subspaces, one for modeling visual information, which is expected to be stable during translation, and the other for capturing the style feature specific to the image.
Intuitively, this can be considered as an extreme case for domain-level semantic decomposition, where each image is considered as a separate domain.

\begin{figure}[t]
\begin{center}
\includegraphics[width=0.95\linewidth]{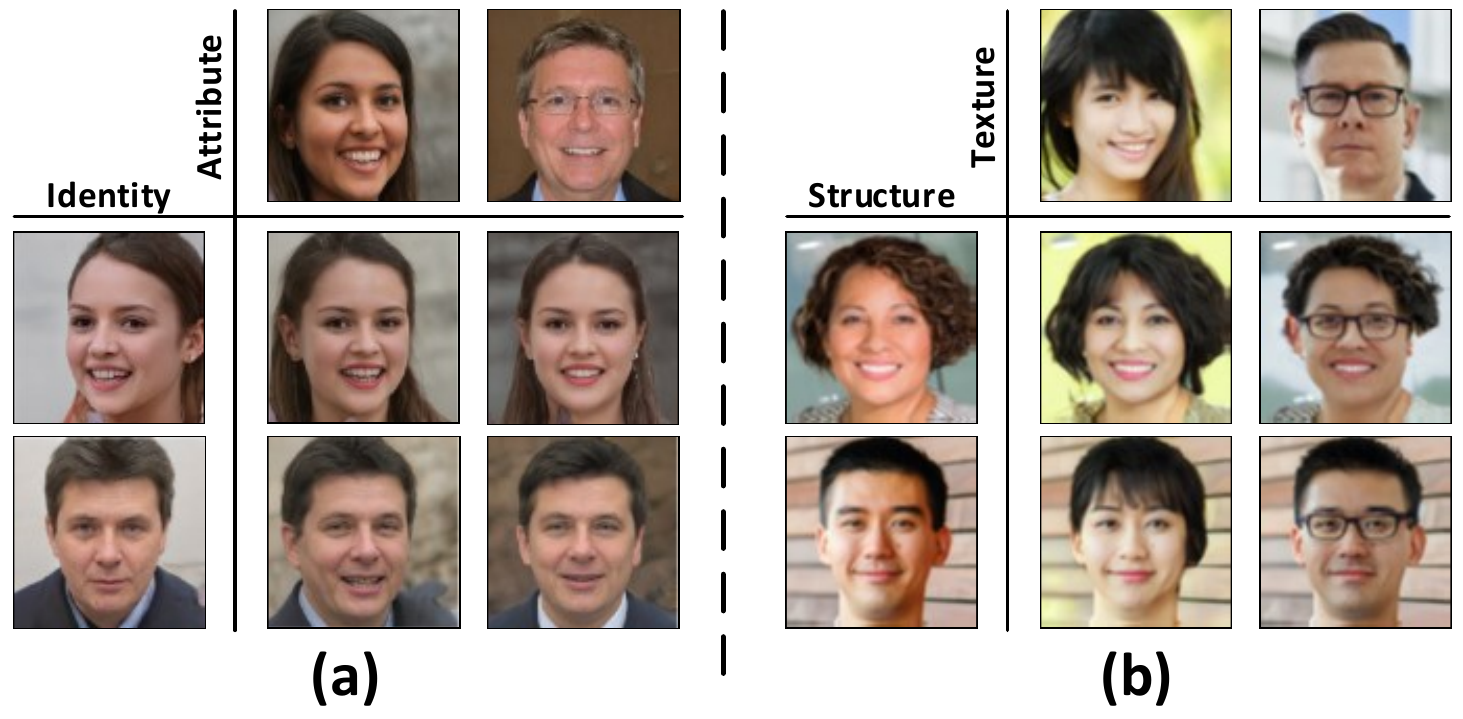}
\end{center}
\vspace{-0.2cm}
\caption{Sample results of (a) LSM~\cite{nitzan2020face} and (b) SwapAE~\cite{park2020swapping}, where source images are shown on the left and exemplar images in the top.
}
\vspace{-0.2cm}
\label{fig:LSM_SwapAE}
\end{figure}

\textnormal{D$\mathbf{^2}$AE}~\cite{liu2018exploring} encodes the identity information and other facial semantics (i.e., attributes) with two separate branches, and the disentanglement is achieved by adversarially training the encoders against an identity classifier (i.e., domain discriminator as in~\cite{lee2018diverse,lee2020drit++}).
Without using adversarial learning, \textnormal{LORD}~\cite{gabbay2019demystifying} learns a disentangled representation via latent optimization, and \textnormal{OverLORD}~\cite{gabbay2021scaling} further analyzes the correlation between labeled and unlabeled facial semantics.
In addition, \textnormal{LSM}~\cite{nitzan2020face} solves a similar problem with the aid of the highly disentangled latent space of a pre-trained StyleGAN generator (see sample results in Fig.~\ref{fig:LSM_SwapAE}(a)).
As shown in Fig.~\ref{fig:LSM_SwapAE}(b), \textnormal{SwapAE}~\cite{park2020swapping}  preserves the overall structure of a face image, i.e., the shape and layout of facial components, and manipulates texture information according to exemplar images.
The consistency in style between FAM results and exemplar images is achieved by aligning the statistic of patch co-occurrence, which is measured by an auxiliary discriminator.

Instead of low-dimensional latent codes,
SensoriumIn~\cite{nederhood2021harnessing} 
operates on the geometry of a face image, which is described by various interpretable 2D representations inherently disentangled with facial textures (e.g., the feature map of facial landmarks), for synthesis. 
\textnormal{SofGAN}~\cite{chen2022sofgan} proposes a semantic occupancy field to render parsed maps with arbitrary viewpoints, which are responsible for controlling the geometry of synthesized images.

Another line of research work disentangles content and attribute by introducing inductive bias into the structure of style-based generators.
\textnormal{SNI}~\cite{alharbi2020disentangled} propose to condition input tensor $\boldsymbol{c}$ on another latent code $\boldsymbol{z_c}$ spatial content manipulation, which is disentangled against the style information controlled by the original latent code $\boldsymbol{z}$.
In~\cite{kwon2021diagonal}, \textnormal{DAT} further makes $\boldsymbol{z_c}$ to have a symmetric structure similar to $\boldsymbol{z}$ and controls the feature map at each scale.
Recently, \textnormal{TransEditor}~\cite{xu2022transeditor} incorporates transformer blocks to establish the interaction between two latent spaces, which improves the controllability and flexibility of FAM.

\subsubsection{Multiple Semantic Decomposition}\label{sec:multi_seman_decomp}

%
Numerous methods are developed to encode face images into multiple latent components responsible for better controlling different facial attributes based on \textnormal{contrastive training batch}, \textnormal{3D graphics model}, and \textnormal{face parsing map}.

\vspace{1mm}
\noindent \textbf{Contrastive Training Batch.}
Numerous methods aim to explicitly associate latent components with different facial attributes, where the disentanglement is achieved by contrasting image samples in a designed training batch.
Given a pair of face images $\boldsymbol{x}$ and $\boldsymbol{y}$ with the opposite label for the $i$-th attribute, \textnormal{DNA-GAN}~\cite{xiao2018dna} uses an encoder $E$ to divide their latent embeddings into multiple segments by 
\begin{align}
&\boldsymbol{z_x}=E(\boldsymbol{x})=[\boldsymbol{x_1},\ldots,\boldsymbol{x_i},\ldots,\boldsymbol{x_n},\boldsymbol{r_x}] \label{eq:DAN_GAN_x_code}\\
&\boldsymbol{z_y}=E(\boldsymbol{y})=[\boldsymbol{y_1},\ldots,\boldsymbol{y_i},\ldots,\boldsymbol{y_n},\boldsymbol{r_y}] \label{eq:DAN_GAN_y_code}
\end{align}
where $\boldsymbol{r_x}$ and $\boldsymbol{r_y}$ are designed to capture the attribute-irrelevant information in $\boldsymbol{x}$ and $\boldsymbol{y}$.
To associate $\boldsymbol{x_i}$ and $\boldsymbol{y_i}$ with the $i$-th attribute ($\boldsymbol{i}=1,2,\ldots,n$), a contrastive training batch containing four latent codes, $\boldsymbol{x}^\ast$, $\boldsymbol{y}^\ast$, $\boldsymbol{x}'$, and $\boldsymbol{y}'$, is obtained by 
\begin{align}
&\boldsymbol{x}^\ast=G([\boldsymbol{x_1},\ldots,\boldsymbol{x_i},\ldots,\boldsymbol{x_n},\boldsymbol{i_x}]), \boldsymbol{y}^\ast=G([\boldsymbol{y_1},\ldots,\boldsymbol{0_i},\ldots,\boldsymbol{y_n},\boldsymbol{i_y}]) \label{eq:DAN_GAN_x_1_y_1}\\
&\boldsymbol{x}'=G([\boldsymbol{x_1},\ldots,\boldsymbol{0_i},\ldots,\boldsymbol{x_n},\boldsymbol{i_x}]), \boldsymbol{y}'=G([\boldsymbol{y_1},\ldots,\boldsymbol{x_i},\ldots,\boldsymbol{y_n},\boldsymbol{i_y}]) \label{eq:DAN_GAN_x_2_y_2}
\end{align}
With this formulation, $\boldsymbol{x}^\ast$ / $\boldsymbol{y}^\ast$ are enforced to approximate $\boldsymbol{x}$ / $\boldsymbol{y}$, and $\boldsymbol{x}'$ / $\boldsymbol{y}'$ are trained to have label 0 / 1 for the $i$-th attribute, respectively.
Thus, the presence/absence of the $i$-th attribute can be achieved by changing the $i$-th component of $\boldsymbol{z}$ to $\boldsymbol{x_i}$/$\boldsymbol{0_i}$.
\textnormal{ELEGANT}~\cite{xiao2018elegant} further improves DNA-GAN by making $G$ to only estimate the residual images against $\boldsymbol{x}$ and $\boldsymbol{y}$, and thus $\boldsymbol{r_x}$ and $\boldsymbol{r_y}$ can be discarded since attribute-irrelevant information is no longer necessary to be modeled.
In addition, \textnormal{GAN-Control}~\cite{shoshan2021gan} also constructs training batches to contain pairs of latent vectors. 
Unlike DNA-GAN and ELEGANT, paired latent codes in GAN-Control share only the $i$-th component and differ in all the rest parts.
Thus, only the $i$-th attribute of corresponding images should be the same, and all other semantics should be different, which is enforced by a contrastive loss.

\vspace{1mm}
\noindent \textbf{3D Graphics Models.}
Apart from carefully designed training batches with paired images, numerous approaches leverage the inherently independent rigging parameters in 3D graphics models for learning disentangled semantic control.
Similar to GAN-Control, \textnormal{DiscoFaceGAN}~\cite{deng2020disentangled} divides the latent space of a pre-trained StyleGAN generator $G^\ast$ into five subspaces, where the latent embeddings (denoted as $\{\boldsymbol{z_i}\}_{i=1}^5$) are responsible for controlling the shape, expression, pose, illumination, and all other textural details of the generated face $G^\ast(\boldsymbol{z})$ ($\boldsymbol{z}=[\boldsymbol{z_1}, \boldsymbol{z_2}, \boldsymbol{z_3}, \boldsymbol{z_4}, \boldsymbol{z_5}]$), respectively.
To supervise the attribute of $G^\ast(\boldsymbol{z})$, a proxy face $\boldsymbol{p}(\boldsymbol{z})$ is rendered using 3DMM, and $G^\ast(\boldsymbol{z})$ is enforced to imitate $\boldsymbol{p}(\boldsymbol{z})$ in terms of the identity, facial landmark layout, skin color, and lighting condition.
The disentanglement among $\{\boldsymbol{z_i}\}_{i=1}^5$ is again achieved by adopting a contrastive loss, which limits the change of image content caused by modifying $\boldsymbol{z_i}$ to the $i$-th attribute.
Similar to DiscoFaceGAN, \textnormal{GIF}~\cite{ghosh2020gif} use dense feature maps obtained by 3DMM for controlling the facial geometry, while the disentanglement is achieved by ensuring the consistency of facial texture across different poses and expressions for the same identity.
In addition, \textnormal{VariTex}~\cite{buhler2021varitex} further models the appearance with a variational latent code, which can be divided into two halves, one for modeling the identity information and the other for the hair and mouth interior.

Instead of 3DMM, \textnormal{ConfigNet}~\cite{kowalski2020config} creates a set of face images with known semantic parameters and finer textural details following~\cite{baltrusaitis2020high}, and trains a generator $G$ on SynthFace to model the mapping between semantic parameters and face images.
An encoder network is then trained to embed real images into the parameter space of $G$, which enables real image manipulation.

\begin{figure}[t]
\begin{center}
\includegraphics[width=0.95\linewidth]{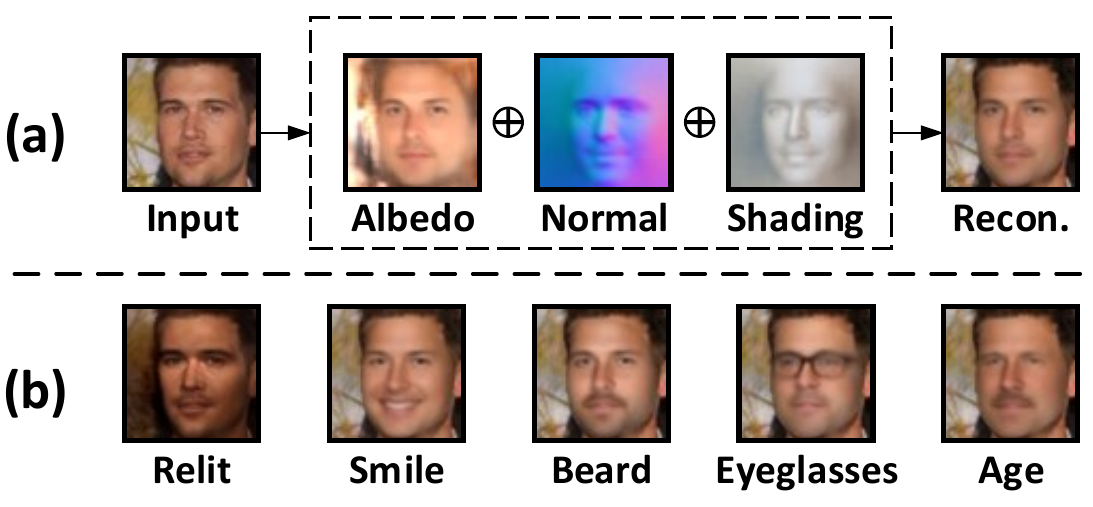}
\end{center}
\vspace{-0.2cm}
\caption{Sample results of NEFNet~\cite{shu2017neural}(a) results of semantic component decomposition and reconstruction (b) FAM results obtained by editing the corresponding latent components.
}
\vspace{-0.2cm}
\label{fig:NEFNet}
\end{figure}

In addition to 3D parametric models, physical face rendering models can also be used for facial semantic factorization.
\textnormal{NFENet}~\cite{shu2017neural} uses the Lambertian rendering model~\cite{barron2014shape} to decompose face images into three physically-based disentangled components, i.e., shape (face normals), albedo, and lighting, which can serve as tuning parameters for facial semantics (as shown in Fig.~\ref{fig:NEFNet}).
Recently, \textnormal{MOST-GAN}~\cite{medin2022most} adopts a similar framework as~\cite{shu2017neural} and proposes an iterative algorithm to achieve fine-grained manipulation of hair regions.

\vspace*{1mm}
\noindent \textbf{Face Parsing Maps.}
In addition to parametric models which intrinsically factorize face images based on high-level semantics, face parsing maps are also widely used to encode spatially disentangled image content with separate latent spaces.
Moreover, since parsing maps contain pixel-wise annotations of facial components, they can be regarded as the representations of image layouts, which naturally allow flexible manipulation.

\begin{figure}[t]
\begin{center}
\includegraphics[width=0.95\linewidth]{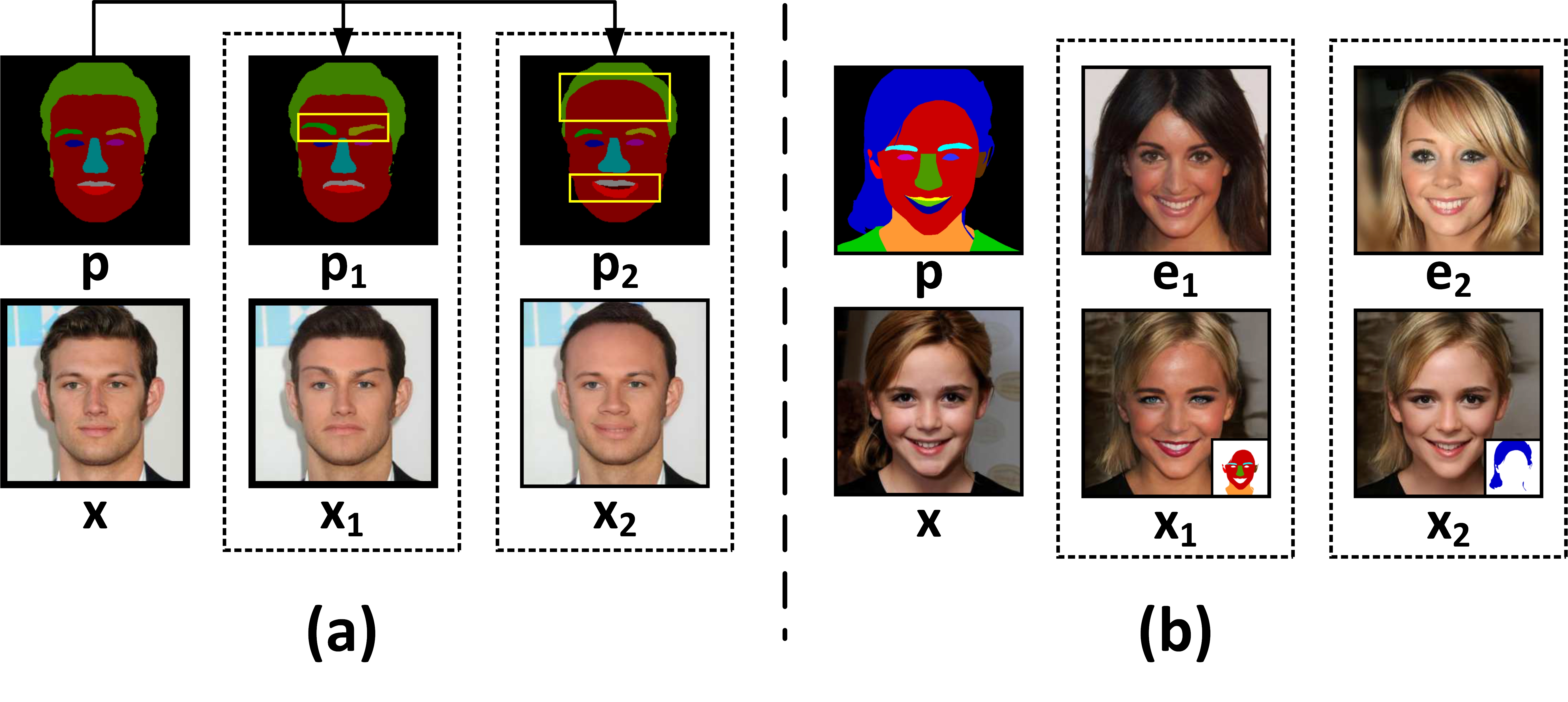}
\end{center}
\vspace{-0.2cm}
\caption{Sample results of (a) MaskGuidedGAN~\cite{gu2019mask} and (b) SEAN~\cite{zhu2020sean}. 
(a) the parsing map $\boldsymbol{p}$ of input image $\boldsymbol{x}$ is modified to $\boldsymbol{p_1}$ and $\boldsymbol{p_2}$ for controlling the facial geometry in $\boldsymbol{x_1}$ and $\boldsymbol{x_2}$ (Yellow boxes highlight the target area for editing.).
(b) $\boldsymbol{x_1}$ and $\boldsymbol{x_2}$ are FAM results of $\boldsymbol{x}$, where the target facial components (indicated by the parsing map at the low-right corner) are modified according to the style of $\boldsymbol{e_1}$ and $\boldsymbol{e_2}$. respectively.
}
\vspace{-0.2cm}
\label{fig:MaskGuidedGAN_SEAN_results}
\end{figure}

With the aid of parsing map $\boldsymbol{p}$, \textnormal{MaskGuidedGAN}~\cite{gu2019mask} decomposes an input image $\boldsymbol{x}$ into five local facial components (i.e., left eye, right eye, mouth, skin \& nose, and hair), and models each one with a separate auto-encoder network.
In the testing phase, the latent representation of these objects can be combined with an arbitrary user-edited parsing map $\boldsymbol{p}'$ to render the corresponding face image, where the geometry and style are determined by $\boldsymbol{p}'$ and $\boldsymbol{x}$, respectively (see Fig.~\ref{fig:MaskGuidedGAN_SEAN_results}(a)).
\textnormal{MaskGAN}~\cite{lee2020maskgan} simplifies MaskGuidedGAN by computing a joint representation for the entire face instead of separately for each facial component, and an \textnormal{Editing Behavior Simulated Training} strategy is proposed to simulate the user editing behavior on parsing maps.
In~\cite{shi2022semanticstylegan}, \textnormal{SemanticStyleGAN} uses a set of local generators to achieve spatially disentangled and semantically compositional face generation and manipulation, where parsing maps are only used for pixel-wise constraints.

Aside from manipulation of each facial component, parsing maps can also be used with exemplar images to achieve fine-grained manipulation of low-level textural details.
Unlike MaskGuidedGAN, \textnormal{SEAN}~\cite{zhu2020sean} uses a single encoder network to compute the style matrix ($\text{ST}$) for input image $\boldsymbol{x}$ via a self-reconstruction process, where each element in $\text{ST}$ encodes the appearance information of an image region indicated by $\boldsymbol{p}$.
In the inference stage, elements in $\text{ST}$ can be replaced by the counterpart of an exemplar image $\boldsymbol{e}$ to achieve controlled FAM, as shown in Fig.~\ref{fig:MaskGuidedGAN_SEAN_results}(b).

\begin{table*}[t]
\caption{Summary of the characteristics of latent space navigation based FAM Methods. Direction/Trajectory Computation indicates how the traversal direction or trajectory is computed. Latent Space shows in which latent space the navigation is performed, where $Z$, $W$, $W^+$, and $S$ refer to latent spaces in style-based generators, and $Z_{SN}$, $Z_{PG}$ and $Z_{Big}$ denote the latent space of SNGAN, PGGAN, and BigGAN, respectively.}
\vspace{-0.2cm}
\rowcolors{1}{}{lightgray}
\centering
\ra{1.3}
\resizebox{2.0\columnwidth}{!}{%
\begin{tabular}{cccccc}
\hline
Method Name & Publication & Navigation Type & Latent Space & Resolution & Quantitative Metrics \\
\hline
StyleGAN2Distillation~\cite{viazovetskyi2020stylegan2} & ECCV 2020  & Linear Interpolation & $W$, $W^+$ & $1024\times 1024$ & FID, US \\
StyleSpaceAnalysis~\cite{wu2021stylespace} & CVPR 2021  & Linear Interpolation & $S$   & $1024\times 1024$ & FID, TARR, DCI~\cite{eastwood2018framework}, Attribute Dependency (AD) \\
InterFaceGAN~\cite{shen2020interpreting}   & CVPR 2020  & Linear Interpolation & $Z_{PG}$, $Z$, $W$ & $1024\times 1024$ & Correlation of Attribute Distributions \\
ACU\cite{wang2021attribute}  & ACM MM 2021  & Linear Interpolation & $S$ & $1024\times 1024$ & FID, AD~\cite{wu2021stylespace}, Success Rate of Local Editing, Region Purity \\
AdvStyle~\cite{yang2021discovering}     & CVPR 2021    & Linear Interpolation & $W$   & $1024\times 1024$ & Correlation of Attribute Distributions \\
EditGAN~\cite{ling2021editgan}          & NeurIPS 2021 & Linear Interpolation & $W^+$ & $1024\times 1024$ & FID, KID, TARR, CSIM \\
EnjoyEditingGAN~\cite{zhuang2021enjoy}  & ICLR 2021    & Linear Interpolation & $Z_{PG}$, $W$ & $1024\times 1024$ & NAPR, CSIM, US \\
Latent-Transformer~\cite{yao2021latent} & ICCV 2021    & Linear Interpolation & $W^+$ & $1024\times 1024$ & The Relation between NAPR/CSIM and Attribute Change \\
Style-Transformer~\cite{hu2022style}    & CVPR 2022    & Linear Interpolation & $W^+$ & $1024\times 1024$ & FID, LPIPS, AD~\cite{wu2021stylespace}, SWD~\cite{rabin2011wasserstein}, Cost Analysis \\
UDID~\cite{voynov2020unsupervised}      & ICML 2020    & Linear Interpolation & $Z_{SN}$, $Z_{PG}$, $Z_{Big}$ & $1024\times 1024$ & TARR, US \\
WarpedGANSpace~\cite{tzelepis2021warpedganspace} & ICCV 2021 & Linear Interpolation & $Z_{SN}$, $Z_{PG}$, $Z_{Big}$, $Z$ & $1024\times 1024$ & TARR, L1-normalized Correlation of Attribute Distributions \\
GANSpace~\cite{harkonen2020ganspace}    & NeurIPS 2020 & Linear Interpolation & $Z_{Big}$, $W$ & $1024\times 1024$ & - \\
SeFa~\cite{shen2021closed}              & CVPR 2021    & Linear Interpolation & $Z_{PG}$, $Z_{Big}$, $Z$ & $1024\times 1024$ & FID, US, Attribute Re-scoring Analysis \\
LowRankGAN~\cite{zhu2021low}            & NeurIPS 2021 & Linear Interpolation & $Z_{Big}$, $Z$ & $1024\times 1024$ & FID, Masked L2 Error of Pixel Value, US, SWD~\cite{rabin2011wasserstein} \\
LatentCLR~\cite{yuksel2021latentclr}    & ICCV 2021    & Linear Interpolation & $Z_{Big}$, $Z$ & $512\times 512$   & US, Attribute Re-scoring Analysis \\
NeuralODE~\cite{khrulkov2021latent}     & ICCV 2021    & Non-linear Traversal & $W$ & $256\times 256$   & US,  Control-Disentanglement Curve \\
SGF~\cite{li2021surrogate}              & CVPR 2021    & Non-linear Traversal & $Z_{PG}$, $W$ & -                 & US, Manipulation Disentanglement Curve/Score \\
HijackGAN~\cite{wang2021hijack}         & CVPR 2021    & Non-linear Traversal & $Z_{PG}$, $Z$ & $1024\times 1024$ & NAPR, modified PPL, Boundary Approximation, Attribute Correlation \\
IALS~\cite{han2021disentangled}         & IJCAI 2021   & Non-linear Traversal & $W$ & $1024\times 1024$ & US, Disentanglement-Transformation Curve \\
StyleRig~\cite{tewari2020stylerig}      & CVPR 2020    & Non-linear Traversal & $W^+$ & $1024\times 1024$ & Change of Latent Component at Different Scales \\
PIE~\cite{tewari2020pie}                & TOG 2020     & Non-linear Traversal & $W$, $W^+$ & $1024\times 1024$ & L2 Error of Head Pose, CSIM, PSNR, SSIM \\
StyleFlow~\cite{abdal2021styleflow}     & TOG 2021     & Non-linear Traversal & $W^+$ & $1024\times 1024$ & FID, LPIPS, CSIM, L1 Error of Attribute Edit Consistency \\
LACE-ODE~\cite{nie2021controllable}     & NeurIPS 2021 & Non-linear Traversal & $W$ & $1024\times 1024$ & FID, TARR, CSIM, Disentangled Edit Strength, Inference Time \\
\hline
\end{tabular}
}
\label{table:summary_latent_space_methods}
\end{table*}

\vspace*{0.1cm}
\noindent \textbf{Unsupervised Factorization.}
Apart from the aforementioned supervised methods, numerous unsupervised methods have been proposed to discover disentangled semantic variations for representation learning. 
\textnormal{InfoGAN}~\cite{chen2016infogan} uses the latent component $\boldsymbol{c}$ to control meaningful semantics by maximizing the mutual information between the distribution of $\boldsymbol{c}$ and generated images $G(\boldsymbol{z},\boldsymbol{c})$.
In~\cite{higgins2017beta}, \textnormal{$\boldsymbol{\beta}$-VAE} impose constraints to limit the capacity of latent factors, such that the model can learn efficient representations of data, i.e., principal semantic variations (facial attributes).
Based on \textnormal{$\boldsymbol{\beta}$-VAE}, \textnormal{FactorVAE}~\cite{kim2018disentangling} achieves a trade-off between semantic disentanglement and reconstruction accuracy by encouraging the marginal distribution of latent representations to be factorial.
Recently, \textnormal{HoloGAN}~\cite{nguyen2019hologan} focuses on learning 3D representations by explicitly incorporating 3D convolutions, rigid-body transformations, and projection units in the generator network.

Instead of learning disentangled latent factors, another line of work decomposes input images in the spatial dimension in an unsupervised manner and then solve for semantic controls for each region.
\textnormal{SCM}~\cite{chen2019semantic} designs a generator with a multi-branch bottleneck, where each branch is responsible for modifying one spatially disentangled facial region along a certain semantic direction.
On the other hand, \textnormal{EIS}~\cite{collins2020editing} adopts a pre-trained StyleGAN generator and factorizes feature maps into independent parts with different semantic meanings via deep feature factorization (DFF).
The latent component responsible for synthesizing the target image region is then located via an optimization algorithm, and FAM can be achieved by modifying it according to the corresponding code of an exemplar image.
\textnormal{StyleFusion}~\cite{kafri2022stylefusion} proposes a hierarchical fusion strategy to disentangle latent representations according to the decomposition patterns obtained by EIS.
In addition,  \textnormal{RIS}~\cite{chong2021retrieve} extends EIS to manipulate more challenging facial attributes which involve large geometric changes (e.g., hairstyle and pose).

\section{Operation on Latent Space}
\label{sec:LSN_FAM}


This class of methods aims to manipulate the attributes of an image $\boldsymbol{x}$ by operating its latent code $\boldsymbol{z_0}$\footnote{For StyleGAN models~\cite{karras2019style,karras2020analyzing,karras2021alias}, we use the $Z$ space for demonstrating operations in the latent space, and similar operations apply for the $W$, $W^+$, and $S$ spaces. 
The notation of equivalent concepts in different spaces share the same formulation and are distinguished by the notation used, e.g., $\boldsymbol{z_0}$/$\boldsymbol{w_0}$/$\boldsymbol{w^+_0}$/$\boldsymbol{s_0}$ refers to the GAN inversion result in the $Z$/$W$/$W^+$/$S$ space, respectively.
For other GAN models, e.g., SNGAN~\cite{miyato2018spectral} (Spectral Norm GAN), BigGAN~\cite{brock2018large} and PGGAN~\cite{karras2017progressive}, only one latent space (i.e., the input latent space referred to as $Z$), is analyzed.} of a pre-trained generator $G^\ast$, such that the resulting image $\boldsymbol{x}'=G^\ast(\boldsymbol{z}')$ can exhibit desired facial attributes.
The edited latent code $\boldsymbol{z}'$ can obtained by \textnormal{linear interpolation} and \textnormal{non-linear traversal}.

\begin{figure}[t]
\begin{center}
\includegraphics[width=0.90\linewidth]{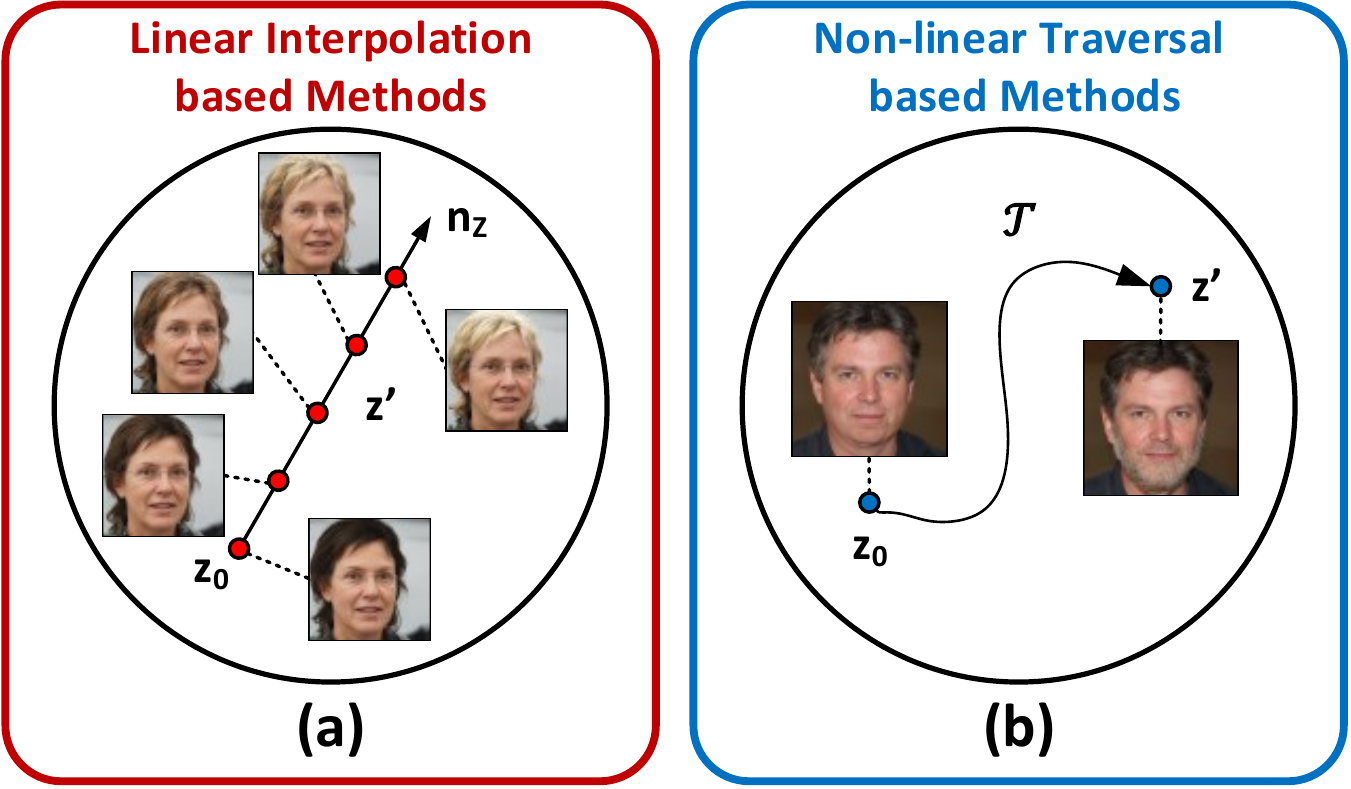}
\end{center}
\vspace{-0.2cm}
\caption{Illustration of latent space navigation based on (a) linear interpolation along the traversal direction $\boldsymbol{n_Z}$, and (b) traveling along a non-linear trajectory $\mathcal{T}$ across the latent space. The target attribute in (a) and (b) are Blonde\_Hair and Beard, respectively.
}
\vspace{-0.2cm}
\label{fig:inter_vs_optim}
\end{figure}

  \textnormal{Linear interpolation} leverages the disentanglement of $G^\ast$'s latent spaces and models the translation process via simple linear interpolation, i.e., $\boldsymbol{z}'=\boldsymbol{z_0}+\alpha\boldsymbol{n_Z}$ (see Fig.~\ref{fig:inter_vs_optim}(a)), where $\boldsymbol{n_Z}$ is the unit vector of the  direction in $Z$ associated with the target facial attribute and $\alpha$ is the step length.

  \textnormal{Non-linear Traversal} assumes that the latent spaces of $G^\ast$ are non-linear, and proposes to make $\boldsymbol{z_0}$ travel along a complex non-linear trajectory $\mathcal{T}$, as shown in Fig.~\ref{fig:inter_vs_optim}(b). 
  Thus, $\boldsymbol{z}'$ can written as $\boldsymbol{z}'=\mathcal{T}(\boldsymbol{z_0})$ where $\mathcal{T}$ is usually implemented by iterative optimization or deep neural networks.
%
A summary of latent space operation based FAM methods is presented in Table~\ref{table:summary_latent_space_methods}.

\subsection{Linear Interpolation}
The focus of linear interpolation based methods is finding the semantically meaningful direction $\boldsymbol{n_Z}$ corresponding to target attributes, with or without supervision. 
%

\subsubsection{Supervised Approaches}

Supervised FAM approaches typically assume that the traversal direction $\boldsymbol{n_Z}$ is uniquely determined by the target attributes and property of $G^\ast$'s latent space but independent of the input latent code $\boldsymbol{z_0}$.
Thus, $\boldsymbol{n_Z}$ can be solved solely based on a large set of synthetic data, which is easy to collect due to the ability of $G^\ast$ in generated realistic images, and then applied to $\boldsymbol{z_0}$ to obtain FAM results.

\begin{figure}[t]
\begin{center}
\includegraphics[width=0.90\linewidth]{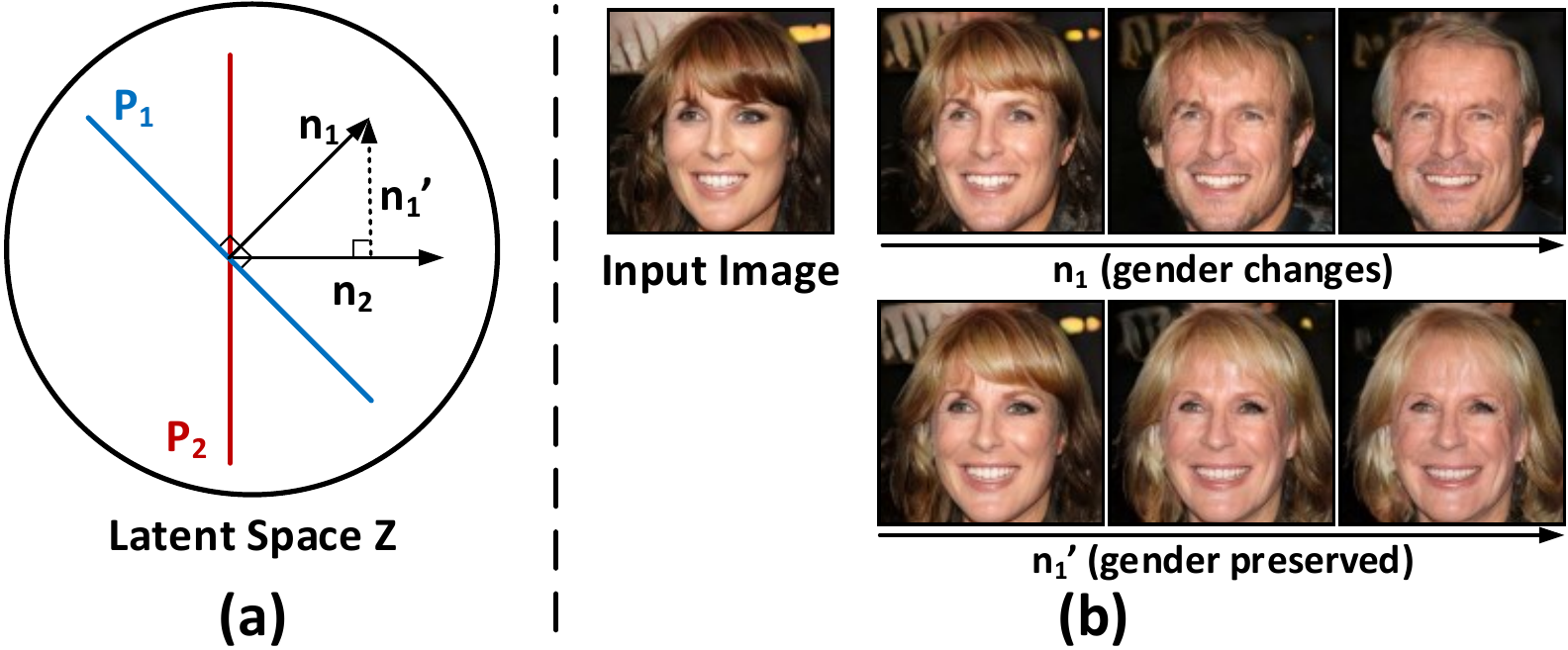}
\end{center}
\vspace{-0.2cm}
\caption{(a) Traversal direction orthogonalization in $Z$, where $\boldsymbol{P_1}$ and $\boldsymbol{P_2}$ are separating hyper-planes of linear SVMs for Age and Gender, and $\boldsymbol{n_1}$ and $\boldsymbol{n_2}$ are the corresponding normal vectors. $\boldsymbol{n_1'}=\boldsymbol{n_1}-(\boldsymbol{n_1}^{\top}\boldsymbol{n_1})\boldsymbol{n_2}$ is the result of orthogonalization. (b) Sample results of manipulating Age, where interpolating along $\boldsymbol{n_1'}$ can eliminate the influence of $\boldsymbol{n_2}$ (Gender) and produce more semantically disentangled FAM results.
}
\vspace{-0.2cm}
\label{fig:InterFaceGAN}
\end{figure}

\textnormal{StyleGAN2Distillation}~\cite{viazovetskyi2020stylegan2} generates synthetic face datasets by randomly sampling latent codes in $Z$, mapping to $W$ (denoted as $\{\boldsymbol{w}_i\}_{i=1}^n$), and synthesizing the corresponding images using the StyleGAN2 generator~\cite{karras2020analyzing}.
Afterwards, the attribute labels $c(\boldsymbol{w}_i)=f(G^\ast(\boldsymbol{w}_i))$ are obtained by a pre-trained facial attribute classifier $f$.
In this formulation, the mean vector $\boldsymbol{c}_k=\frac{1}{n_k}\sum_{c(\boldsymbol{w}_i)=k}\boldsymbol{w}_i$ can be regarded as representative of images with the $k$-th attribute, and thus the difference $\boldsymbol{n_W}_{i,j}\triangleq\boldsymbol{c}_j-\boldsymbol{c}_i$ denotes the traversal direction from the $i$-th class to the $j$-th class in $W$.
In \textnormal{StyleSpaceAnalysis}~\cite{wu2021stylespace}, the translation vector is computed in the $S$ space for better disentanglement, and much fewer samples are required (e.g., 10-30) to compute $\boldsymbol{n_S}$.
\textnormal{ACU}~\cite{wang2021attribute} extends StyleSpaceAnalysis by also manipulating feature maps in the generator, which can achieve more realistic FAM results without damaging the spatial disentanglement of image changes.
On the other hand, 
\textnormal{InterFaceGAN}~\cite{shen2020interpreting} models $\boldsymbol{n_Z}$ as the normal vector of hyper-planes separating latent codes corresponding to different attribute labels.
It improves the disentanglement between two traversal directions $\boldsymbol{n_1}$ and $\boldsymbol{n_2}$ via orthogonalization, i.e., subtracting the projection of $\boldsymbol{n_1}$ onto $\boldsymbol{n_2}$ from $\boldsymbol{n_1}$, as shown in Fig.~\ref{fig:InterFaceGAN}.

\begin{figure}[t]
\begin{center}
\includegraphics[width=0.90\linewidth]{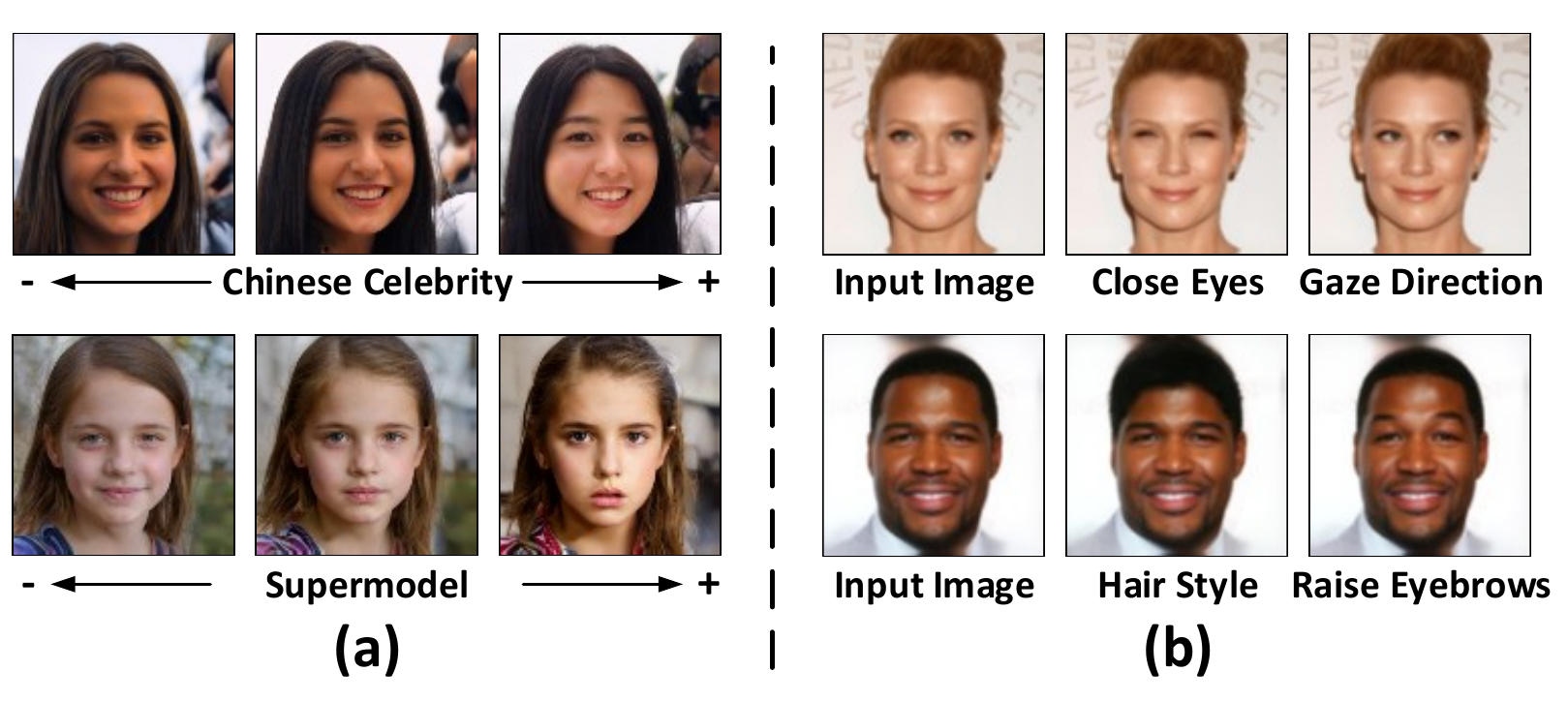}
\end{center}
\vspace{-0.2cm}
\caption{Sample FAM results obtained by (a) AdvStyle~\cite{yang2021discovering} (input images are in the middle) and (b) EditGAN~\cite{ling2021editgan} (results obtained with latent editing directions learned on user-edited samples). Target facial attributes are labeled underneath images.
}
\vspace{-0.2cm}
\label{fig:AdvStyle_EditGAN}
\end{figure}

Apart from attribute labels obtained by pre-trained classifiers, various annotation schemes are proposed to supervise the effect of latent navigation.
To manipulate novel facial attributes (e.g., Supermodel and Chinese Celebrity, see Fig.~\ref{fig:AdvStyle_EditGAN}(a)) beyond well-established ones with binary labels, \textnormal{AdvStyle}~\cite{yang2021discovering} introduces a deep model to discriminate the target attribute from transformed images, which is simultaneously updated with $\boldsymbol{n_W}$.
To perform user-defined geometric changes of facial components (as shown in Fig.~\ref{fig:AdvStyle_EditGAN}(b)), \textnormal{EditGAN}~\cite{ling2021editgan} allows users to edit the parsed map of an input image and finds the latent displacement vector to reflect such modification via an optimization-based method.
Due to the structured and disentangled latent space of $G^\ast$, the editing vector learned from one input sample can be generalized to other images for performing semantically similar manipulations.

However, some studies point out that the latent space of $G^\ast$ is not guaranteed to be uniform, and thus $\boldsymbol{n_Z}$ is not globally applicable and should be dependent on the source latent embedding, i.e. $\boldsymbol{n_Z}=\boldsymbol{n_Z}(\boldsymbol{z_0})$.
\textnormal{EnjoyEditingGAN}~\cite{zhuang2021enjoy} uses a simple multilayer perceptron (MLP) to compute the translation vector based on $\boldsymbol{z_0}$ for manipulating one facial attribute.
As it learns multiple traversal directions for FAM simultaneously, it is likely to disentangle and control image changes better.
A binary cross entropy loss is adopted for supervision, which associates the step length of interpolation with the change of regressed attribute scores. %
\textnormal{Latent-Transformer}~\cite{yao2021latent} uses a transformer in $W^+$ to predict the traversing direction $\boldsymbol{n_{W^+}}$ based on the embedding of the input images $\boldsymbol{w}^+_0$.
In addition, transformer blocks are also used in \textnormal{Style-Transformer}~\cite{hu2022style} to achieve flexible FAM based on both attributes labels and exemplar images.

\subsubsection{Unsupervised Approaches}
To reduce the annotation cost of training data, numerous unsupervised methods have been proposed to discover interpretable semantic directions $\boldsymbol{n_Z}$ for FAM.
The key is to describe the properties of the desired $\boldsymbol{n_Z}$ with a concrete objective function, which can be solved via optimization.

Without any form of supervision, \textnormal{UDID}~\cite{voynov2020unsupervised} discovers a set of traversal directions where the induced image transformations could be distinguished from each other.
Specifically, an auxiliary regression network $R$ is adopted to predict $\boldsymbol{n_Z}$ and the $\alpha$ from a pair of images $G(\boldsymbol{z})$ and $G(\boldsymbol{z}+\alpha\boldsymbol{n_Z})$, which is optimized jointly with $\boldsymbol{n_Z}$ in a self-supervised manner.
\textnormal{WarpedGANSpace}~\cite{tzelepis2021warpedganspace} adopts a framework similar to UDID with $\boldsymbol{n_Z}$ estimated by a set of Radial basis functions (RBFs).
On the other hand, \textnormal{GANSpace}~\cite{harkonen2020ganspace} models interpretable traversal directions as the principal components of feature tensors in the $W$ space, which capture the major semantic variations of training data (i.e., facial attributes) and can be solved by Principal Components Analysis.
Recently, \textnormal{SeFa}~\cite{shen2021closed} models the mapping between primitive latent codes $\boldsymbol{z}\in Z$ and intermediate features $\boldsymbol{w}\in W$ as an affine transformation, i.e., $\boldsymbol{w}=f(\boldsymbol{z})=\boldsymbol{A}\boldsymbol{z}+\boldsymbol{b}$ ($\boldsymbol{A}$ and $\boldsymbol{b}$ are the weight and bias parameters, respectively), and claims that the change in $\boldsymbol{w}$ should be maximized when traversing along $\boldsymbol{n_Z}$ associated with meaningful semantics.
In addition, \textnormal{LowRankGAN}~\cite{zhu2021low} models the relationship between $\boldsymbol{z}$ and $G^\ast(\boldsymbol{z})$ using the Jacobian matrix $(\boldsymbol{J}_z)_{j,k}=\frac{\partial G^\ast(\boldsymbol{z})_j}{\partial \boldsymbol{z}_k}$, and determines $\boldsymbol{n_Z}$ by maximizing the change of image $G(\boldsymbol{z})$:
\begin{align}
\boldsymbol{n}&=\argmaxA_{\Vert\boldsymbol{n}\Vert_2=1} \Vert G(\boldsymbol{z}+\alpha\boldsymbol{n_Z})-G(\boldsymbol{z})\Vert_2 \\
              &\approx \argmaxA_{\Vert\boldsymbol{n}\Vert_2=1} \alpha^2\boldsymbol{n}^{\top}\boldsymbol{J}_z^{\top}\boldsymbol{J}_z\boldsymbol{n_Z}
\label{eq:lowrankgan}
\end{align}
The solution is obtained via the low-rank factorization of $\boldsymbol{J}_z^{\top}\boldsymbol{J}_z$. 
It also shows that interpolating along the determined direction $\boldsymbol{n_Z}$ can generate spatially disentangled FAM results.
Similar to the idea in previous methods, \textnormal{LatentCLR}~\cite{yuksel2021latentclr} assumes variation of visual appearance caused by modifying $\boldsymbol{z}$ could be identifiable if $\boldsymbol{n_Z}$.
Thus, $\boldsymbol{n_Z}$ can be obtained by imposing contrastive constraints. 

\subsection{Non-linear Traversal}

The trajectories of non-linear traversal can mainly be computed with \textnormal{iterative optimization} or \textnormal{deep neural networks}.

\subsubsection{Iterative Optimization}
Several methods operate on a non-linear latent space by dividing the traversing process into multiple steps, and interpolating within a neighborhood at each step.
%
\textnormal{NeuralODE}~\cite{khrulkov2021latent} parameterizes the traversal in latent space by Neural ODE~\cite{neural2018tian}, and thus linear interpolation could be considered as a special case where the first derivative is a constant, i.e., $\dot{\boldsymbol{w}}=\boldsymbol{n_W}$ (the initial condition is set to $\boldsymbol{w}_0$).
Following this formulation, non-linear latent traversal could be realized by replacing the right-hand side term with $f(\boldsymbol{w})$, where $f$ is usually implemented with deep neural networks.
\textnormal{SGF}~\cite{li2021surrogate} specifies $f$ as the surrogate gradient field determined by the desired property of manipulation results (e.g., facial attributes or landmark layout).

To approximate non-linear latent trajectories, \textnormal{HijackGAN}~\cite{wang2021hijack}  trains a proxy model $P$ to map $\boldsymbol{z}$ to $M\circ{G}^\ast(\boldsymbol{z})$, where $M$ is an attribute classifier.
The row vector of the Jacobian matrix $\boldsymbol{J}$ of $P$, 
$\boldsymbol{J}_{j}=[\frac{\partial P_j}{\partial z_1},\ldots,\frac{\partial P_j}{\partial z_n}]^{\top}$, describes the traversal direction for manipulating the $j$-th attribute ($P_j$ denotes the $j$-th attribute predicted by $P$).
Consequently, $\boldsymbol{z}$ can be iteratively updated by $\boldsymbol{z}^{(t+1)}=\boldsymbol{z}^{(t)}-\alpha \boldsymbol{J}_{j}(\boldsymbol{z}^{(t)})$, where the traversal direction $\boldsymbol{J}_{j}(\boldsymbol{z}^{(t)})$ is dependent on $\boldsymbol{z}^{(t)}$ at step $t$.
A similar iterative update process is proposed in \textnormal{IALS}~\cite{han2021disentangled}, where the traversal direction at each step contains two parts: an instance-specific semantic direction $\boldsymbol{d}_{l}$ (i.e., local direction dependent on $\boldsymbol{z}$) and an attribute-level semantic direction $\boldsymbol{d}_{g}$ (i.e., global direction independent of $\boldsymbol{z}$).
Specifically, $\boldsymbol{d}_{l}$ is determined via gradient descent and can be written as $-\nabla_{\boldsymbol{z}}\mathcal{L}(M\circ{G}^\ast(\boldsymbol{z}))$, where $\mathcal{L}$ is an attribute classification loss, and $\boldsymbol{d}_{g}$ is obtained by averaging $\boldsymbol{d}_{l}$ on a large collection of $\boldsymbol{z}$.

\subsubsection{Deep Neural Networks}
Apart from iterative interpolation-based methods, deep neural networks are also used to determine the target latent code $\boldsymbol{z}'$.
Given the latent embeddings of source and exemplar face images (denoted as $\boldsymbol{w_0}$ and $\boldsymbol{w_e}$, respectively), \textnormal{StyleRig}~\cite{tewari2020stylerig} computes the rigging parameter $\Delta\boldsymbol{w}$ as $\Delta\boldsymbol{w}=D(E_0(\boldsymbol{w_0}), E_e(\boldsymbol{w_e}))$, where $E_0$, $E_e$, and $D$ are implemented using MLP networks.
\textnormal{PIE}~\cite{tewari2020pie}  improves StyleRig by jointly optimizing the image embedding and manipulation modules.
In~\cite{abdal2021styleflow} \textnormal{StyleFlow} uses a bi-directional network, which is constructed with a sequence of continuous normalizing flow blocks, to embed the input image via a joint reverse encoding process and render the FAM results through a conditional forward editing process.
%
On the other hand, \textnormal{LACE-ODE}~\cite{nie2021controllable} represents the joint distribution of data and attributes with an energy-based model (EBM) in the latent space, and thus FAM could be performed by changing the combination of latent attribute codes.

\section{Properties of GAN-based FAM Methods}\label{sec:property}
In this section, we summarize the important properties of GAN-based FAM methods in five aspects, including \textnormal{visual quality}, \textnormal{desired target attribute}, \textnormal{semantic consistency}, \textnormal{diversity and controllability}, and \textnormal{editing multiple features}.

\subsection{Visual Quality of FAM Results}\label{sec:visual_quality}

FAM results should be as photo-realistic as possible in terms of both overall structure and fine-grained details.
The improvement on visual quality and image resolution can largely be attributed to the rapid development of GAN-based image synthesis methods, including training objectives~\cite{goodfellow2014generative,mao2017least,arjovsky2017wasserstein,gulrajani2017improved} and network structures~\cite{radford2016unsupervised,karras2017progressive,karras2019style,karras2020analyzing,karras2021alias}.
These advancements help stabilize the adversarial training process between the generator and discriminator, and eliminate ghost artifacts in synthesized images.
The recent success of style-based GANs~\cite{karras2019style,karras2020analyzing,karras2021alias} in synthesizing HR images and learning disentangled semantic representations has enabled FAM based on latent space navigation~\cite{harkonen2020ganspace,shen2020interpreting,shen2021closed}.
%
Since the parameters of style-based generators in these models do not need to be updated, the computational complexity of these methods is significantly reduced and the synthesized results after manipulation are often photo-realistic. 


\subsection{Desired Target Attributes}\label{sec:translation_fulfillment}

For any FAM method, the modified target attributes should be successfully recognized in synthesized images. 
This goal is usually achieved either by explicitly imposing an attribute classification loss or implicitly associating different latent components with disentangled facial attributes.

Attribute classification loss is usually used in FAM methods based on image domain translation and latent space navigation.
FAM models learn to render facial attributes as desired in translation results by minimizing the attribute classification loss, which is commonly implemented by using an auxiliary recognition network~\cite{he2019attgan,deng2020controllable,yang2021discovering} or multi-tasking the discriminator~\cite{choi2018stargan,liu2019stgan,wang2019sdit}.

For semantic decomposition based FAM methods, desired results are obtained by manipulating the latent components corresponding to target attributes. 
Thus, the key is to closely link each decomposed facial semantic to a designated latent component, which is usually ensured by imposing contrastive losses~\cite{xiao2018dna,xiao2018elegant,shoshan2021gan} or involving parametric 3D graphics models~\cite{deng2020disentangled,piao2021inverting,ghosh2020gif,buhler2021varitex,kowalski2020config,shu2017neural,medin2022most}.

\subsection{Preservation of Facial Semantics}\label{sec:preservation_semantic}

In addition to correctly synthesizing images with desired attributes, other facial regions are also expected to be preserved after manipulation. 
This can be achieved by imposing direct constraints on image consistency or better disentanglement among semantic regions.

Pixel-level constraints are widely used to enforce modifications only on regions closely related to the target attributes, which are either learned during training~\cite{zhang2018generative,tang2021attentiongan} (e.g., via the attention mechanism) or indicated by face parsing maps~\cite{liu2020style}.
Some methods regulate the synthesis process by penalizing the difference of all pixels between input and reconstructed images, which can be obtained by cyclic translation~\cite{zhu2017unpaired,yi2017dualgan,kim2017learning} (i.e., cycle-consistency loss) or identity mapping~\cite{he2019attgan,siddiquee2019learning}.
Moreover, feature-level constraints are also employed by FAM methods to ensure perceptual consistency between input and edited images (e.g. identity preservation~\cite{li2016deep}). 

Apart from explicitly imposing training objectives to preserve visual information, several methods implicitly disentangle the factors for different facial attributes, and thus the manipulation of target attributes can be constrained to proper regions.
Such disentanglement can be achieved by adversarially training with an auxiliary discriminator~\cite{lee2018diverse,lee2020drit++}, contrastive learning with specially designed training batches~\cite{xiao2018dna,xiao2018elegant,shoshan2021gan}, leveraging the intrinsic disentanglement of 3D models~\cite{deng2020disentangled,kowalski2020config,piao2021inverting,ghosh2020gif,buhler2021varitex}, or using face parsing maps for locating individual facial component~\cite{gu2019mask,lee2020maskgan,shi2022semanticstylegan,zhu2020sean,collins2020editing,kafri2022stylefusion,chong2021retrieve}.

\subsection{Diversity and Controllability of Generated Images}\label{sec:diversity_controllability}
Facial attributes typically describe perceptually salient properties of human faces, where no specific restrictions are imposed on the shapes or textures of facial components in manipulated images.
Thus, FAM models can generate diverse (also referred to as multi-modal) outputs.
However, most image domain translation based methods~\cite{zhu2017unpaired,choi2018stargan,he2019attgan} are unimodal, and only inter-domain mapping functions are
estimated
To address this issue, a few  approaches~\cite{wang2019sdit,romero2019smit}  incorporate random variables for modeling intra-domain variation of the target domain, and thus diverse FAM results can be obtained by re-sampling at test time.

To improve the controllability of multi-modal results, some FAM methods use style information~\cite{lee2018diverse,lee2020drit++,choi2020stargan,li2021image}, which encodes the intra-domain variation specific to a certain subject (i.e., exemplar face), for guiding the texture and geometry of translation results.
In particular, a few methods~\cite{lee2018diverse,lee2020drit++} align the distribution of style code with random variables, which makes the model be compatible with two types of data as guidance, i.e., noise vectors and exemplar images.
Multi-modal FAM methods can also be obtained by simply style mixing~\cite{karras2019style,karras2020analyzing} with latent codes sampled from the prior distribution or extracted from exemplar images for better controllability.

\subsection{Editing Multiple Attributes}\label{sec:scalability}

For image domain translation based FAM results, two-domain approaches~\cite{li2016deep,shen2017learning,tang2021attentiongan} can only manipulate one facial attribute with a single model, and thus $2n$ new models need to be trained for every new attribute, where $n$ is the number of attributed features.
The capacity of multi-domain approaches~\cite{hui2018unsupervised,zhao2018modular,zhang2018sparsely} is determined by the number of attribute labels available in training data, and the model has to be re-trained whenever new attributes are introduced for editing.

For semantic decomposition based FAM methods, the number of features that can be modified is largely determined by the internal mechanism for dividing facial semantics.
Specifically, FAM methods trained with contrastive data~\cite{xiao2018dna,xiao2018elegant,shoshan2021gan} can be extended to manipulate multiple features (as long as paired images with the target attribute can be collected), but may be less effective for disentanglement as the number of attributes increases.
However, FAM methods based on parametric 3D models~\cite{deng2020disentangled,kowalski2020config,piao2021inverting,ghosh2020gif,buhler2021varitex} and face parsing maps~\cite{gu2019mask,lee2020maskgan,zhu2020sean}
for explicit semantic decomposition 
cannot be easily extended to handle multiple features as specific prior knowledge need to be incorporated (e.g., number of categories in parsing maps).

FAM methods based on latent space traversal can be easily extended to manipulate multiple features, as the generator pre-trained on unconditional image synthesis has learned to extract the representation of different semantics.
%
Moreover, since the parameters of the generator are fixed, the cost of adapting this class of FAM methods to editing unseen attributes is lower than that of other approaches.

\section{Challenges and Future Directions}\label{sec:challenge_future}

\noindent\textbf{Fine-grained Control of Individual Components.}
Most existing GAN-based FAM methods, conditioned either on label vectors or style codes, can hardly provide low-level controls on the shape and texture of manipulation results.
For instance, fine-grained control of the fine-grained shape and structure of hair, the angle of gaze direction, and head pose, cannot be accomplished when editing the associated attributes by existing approaches.
To solve this problem, some studies incorporate pixel-wise conditional information obtained by user interaction~\cite{portenier2018faceshop,jo2019sc,tan2020michigan,ling2021editgan} as guidance.
If is of great interest to develop automated and generalized solutions that allow fine-grained feature manipulation.

\vspace{0.1cm}
\noindent\textbf{Task-specific Inductive Biases.}
Although general-purpose image translation methods (e.g., CycleGAN~\cite{zhu2017unpaired}) can be used for editing facial attributes, introducing task-specific inductive biases can help further improve the quality of FAM results.
For example, 3D Morphable Models (3DMMs) are widely used for manipulating pose and expression~\cite{deng2020disentangled,piao2021inverting,buhler2021varitex,medin2022most}, as face shape can be factorized into disentangled controlling parameters.
Most recently, Neural Radiance Fields (NeRFs) have been used as an implicit presentation to model fine-grained 3D structure of faces~\cite{schwarz2020graf,chan2021pi,sun2022fenerf,xu20223d}, where the volumetric rendering mechanism is introduced for reconstruction.
Thus, developing mechanisms to introduce more efficient inductive biases specific to the target facial attribute is a promising future research direction.

\vspace*{0.1cm}
\noindent\textbf{GAN Inversion and FAM.}
Latent codes are usually obtained via GAN inversion, and then manipulated to generate desired FAM results (i.e., in two independent steps).
However, the main objective of GAN inversion is to ensure that the latent embedding can faithfully reconstruct an input image, but does not guarantee that it can be edited to render desired effects.
Moreover, some methods~\cite{zhu2020domain} show that the latent codes generated by GAN inversion models may locate outside of the original distribution, and thus severe ghost artifacts can be observed after manipulation.
Thus, GAN inversion and latent code manipulation should be jointly considered for better FAM results, which has recently gained some attention~\cite{richardson2021encoding,dinh2022hyperinverter}.

\vspace*{0.1cm}
\noindent\textbf{Biased Generative Priors.}
Although pre-trained generators are widely used in FAM methods for image synthesis, the distribution of output images is restricted and biased towards the training dataset.
For example, most face images synthesized by StyleGAN generators pre-trained on FFHQ have neutral or smiling expressions with the near-frontal view.
Thus, it is difficult for these networks to model faces with large poses or arbitrary expressions~\cite{zhu2021one}.
It is of great interest to mitigate such biased priors by adapting the domain-agnostic semantics learned by pre-trained GAN models for tasks with different data distributions.

\vspace*{0.1cm}
\noindent\textbf{Guiding Information with Multimodality.}
In most existing GAN-based FAM methods, the target attributes are specified by labels~\cite{he2019attgan,choi2018stargan,choi2020stargan} or exemplar images~\cite{xiao2018elegant,li2021image,zhu2020sean}.
Recently, conditional information in other modalities, such text~\cite{patashnik2021styleclip,xia2021tedigan,jiang2021talk,sun2021multi,sun2022anyface} and speech~\cite{han2022show,guo2021ad,ji2021audio}, has attracted increasing research attention due to the development of pre-trained large-scale frameworks (e.g., CLIP~\cite{radford2021learning}) and availability of related datasets (CelebA-Dialog~\cite{zhu2022celebvhq}).
%
Moreover, novel modalities of supervision signal, such as biometrics (e.g., brain responses recorded via electroencephalography~\cite{davis2022brain}) and sound~\cite{lee2022sound}, have also been utilized to learn feature representations for semantic editing.

\vspace*{0.1cm}
\noindent\textbf{Video-based FAM.}
The recent development in DeepFake~\cite{perov2020deepfacelab,tolosana2020deepfakes} and face re-enactment~\cite{kim2018deep,hong2022depth} have demonstrated great societal impact and  significant practical value of FAM on video data.
Moreover, video manipulation can also serve as a data augmentation method for DeepFake detection methods, and thus plays an important role in information forensics.
Most recently, a new video dataset with facial attribute annotation of emotion, action and appearance is released (CelebV-HQ~\cite{zhu2022celebvhq}), which shows growing interest in video-based FAM.

\section{Conclusions}
\label{sec:conclusion}
Facial attributes intuitively describe the representative properties of human faces, and have received much attention in the field of vision and learning. 
As one of the most widely-used research topics, facial attribute manipulation has both important research and application values. 
This paper presents a comprehensive survey of existing GAN-based FAM studies.
We analyze the similarities and differences between these methods in terms of motivation and technical details.
Moreover, we summarize the important properties of FAM as well as how they are approached by existing methods.
We conclude this survey with discussions of the challenges and future research directions of FAM.

\ifCLASSOPTIONcaptionsoff
  \newpage
\fi



%
\bibliographystyle{IEEEtran}
\bibliography{IEEEbib}

%








\end{document}